\documentclass{article}

\usepackage{PRIMEarxiv}

\usepackage[utf8]{inputenc} 
\usepackage[T1]{fontenc}    
\usepackage{hyperref}       
\usepackage{url}            
\usepackage{booktabs}       
\usepackage{amsfonts}       
\usepackage{nicefrac}       
\usepackage{microtype}      
\usepackage{lipsum}
\usepackage{natbib}
\usepackage{amsmath}
\usepackage[percent]{overpic}
\usepackage{fancyhdr}       
\usepackage{graphicx}       
\graphicspath{{media/}}     
\RequirePackage{graphicx,epic}
\usepackage{dsfont}
\title{Reduced-order modeling for parameterized large-eddy simulations of atmospheric pollutant dispersion
}

\author{
  B.X. Nony\thanks{CECI, Université de Toulouse, CNRS, Cerfacs, 42 Avenue Gaspard Coriolis , 31057 Toulouse cedex 1, France} \\
  \texttt{bastien.nony@cerfacs.fr} \\
   \And
  M.C. Rochoux\footnotemark[1]\\
  \texttt{melanie.rochoux@cerfacs.fr} \\
   \And
  T. Jaravel\footnotemark[1]\\
  \texttt{thomas.jaravel@cerfacs.fr} \\
   \And
  D. Lucor\thanks{LISN, CNRS, Université Paris-Saclay, Campus Universitaire bât 507, Rue John von Neumann, 91405 Orsay cedex, France} \\
  \texttt{didier.lucor@lisn.upsaclay.fr} \\
}

\begin{document}
\maketitle
\begin{abstract}
Mapping near-field pollutant concentration is essential to track accidental toxic plume dispersion in urban areas. By solving a large part of the turbulence spectrum, large-eddy simulations (LES) have the potential to accurately represent pollutant concentration spatial variability. Finding a way to synthesize this large amount of information to improve the accuracy of lower-fidelity operational models (e.g. providing better turbulence closure terms) is particularly appealing. This is a challenge in multi-query contexts, where LES become prohibitively costly to deploy to understand how plume flow and tracer dispersion change with various atmospheric and source parameters. To overcome this issue, we propose a non-intrusive reduced-order model combining proper orthogonal decomposition (POD) and Gaussian process regression (GPR) to predict LES field statistics of interest associated with tracer concentrations. GPR hyperpararameters are optimized component-by-component through a maximum a posteriori (MAP) procedure informed by POD. We provide a detailed analysis of the reduced-order model performance on a two-dimensional case study corresponding to a turbulent atmospheric boundary-layer flow over a surface-mounted obstacle. We show that near-source concentration heterogeneities upstream of the obstacle require a large number of POD modes to be well captured. We also show that the component-by-component optimization allows to capture the range of spatial scales in the POD modes, especially the shorter concentration patterns in the high-order modes. The reduced-order model predictions remain acceptable if the learning database is made of at least fifty to hundred LES snapshot providing a first estimation of the required budget to move towards more realistic atmospheric dispersion applications.
\end{abstract}


\section{Introduction}

Accidental short-term pollutant emissions (e.g. 2011 Fukushima power plant explosion -- \citeauthor{tsuruta2014}, \citeyear{tsuruta2014}; 2019/2020 Australian bushfires -- \citeauthor{graham2021}, \citeyear{graham2021}) can significantly degrade air quality and impact public health. In urban areas, peak pollutant concentrations are difficult to track due to the complex interactions between meteorology and urban topography~\citep{philips2013large}. As a result, predicting the range of possible scenarios for near-source air pollutant dispersion in urban areas is critical for decision support in emergency situations~\citep{da2021impact}.

Computational fluid dynamics (CFD) is a complementary approach to field campaigns (e.g.~Oklahoma City Joint Urban 2003 Experiment -- \citeauthor{allwine2006joint}, \citeyear{allwine2006joint}) to study microscale atmospheric dispersion processes in a complex urban environment~\citep{tominaga2013review,dauxois2021}. CFD has first been tackled through Reynolds-averaged Navier-Stokes (RANS) approaches~\citep{milliez2007,garcia2017},  for which a time-averaged solution to the Navier-Stokes equations is sought, and turbulence is fully modeled via an approximate closure. By explicitly solving the large turbulent scales characterizing the flow, large-eddy simulation (LES) has emerged in recent years as an accurate reference solution to represent spatio-temporal variability of turbulent atmospheric flows in urban environment~\citep{philips2013large,vervecken2015dose,garcia2018les,grylls2019}. LES has a superior prediction capability compared to RANS, as it accurately captures highly unsteady and complex flow topologies typically found in the wake of buildings in urban canopies~\citep{tominaga2010numerical}. It also provides higher order statistics such as concentration fluctuation and turbulent scalar flux fields that are not available with RANS. Yet, LES demands much larger computational cost, which is a bottleneck for application in an uncertainty quantification context.

Many sources of uncertainty in microscale CFD simulations come from large-scale atmospheric conditions that are difficult to represent due to their intrinsic variability, but also from limited knowledge on emission source properties. To deal with these uncertainties and build a multi-query uncertainty quantification context suitable for CFD, one promising approach consists in substituting the parametric CFD model by a reduced-order model. The reduced-order model can be seen as a statistical model that has learned from a reduced CFD database, the relationship between uncertain input parameters and the flow/tracer quantities of interest, and that can then estimate the CFD model response over a wide range of parametric variation~\citep{margheri2016hybrid,garcia2017,lamberti2021}. For example, \citet{garcia2017} studied how atmospheric uncertainties (inlet wind magnitude/direction and ground roughness) propagate on plume dispersion in Oklahoma City based on ensemble RANS simulations combined with polynomial chaos expansion. 

In this work, we aim at designing and evaluating a reduced-order model suitable for LES. One key issue when building a CFD reduced-order model relates to the high dimension of the CFD outputs and in the possible nonlinear input-output relationship that is difficult to learn from a limited training database. This is especially true in a LES setting that requires more computational resources than RANS, implying that the training stage can only rely on a very limited LES database. To make the reduced-order model approach tractable in a LES framework, we adopt a two-step reduced-order model approach. The first step is to reduce the LES output dimension, i.e. by projecting the quantities of interest on a reduced basis using proper orthogonal decomposition (POD)~\citep{margheri2016hybrid,van2017uncertainty}. The second step is to interpolate (or to metamodel) the POD coefficients on the uncertain input parameters. The resulting reduced-order model can then be used to predict the quantities of interest for any input parameter entry. 

Since a reduced basis generally accounts for non-affine input/output relationships, basic interpolation techniques for the POD coefficients mapping may fail unless a sufficient number of LES samples is used during the training stage. In the literature, a wide variety of methods have been applied for predictive regression learning on numerical data such as neural networks~\citep{hesthaven2018non,swischuk2019projection,ma2021,lucor2022}, polynomial chaos expansion~\citep{garcia2014quantifying,garcia2017,garroussi2022}, Gaussian process regression (GPR) models~\citep{margheri2016hybrid,guo2018reduced,xiao2019domain}, and decision trees~\citep{xiang2021non}. The search for adapted machine learning tools coincides with the requirement for a careful and documented study of the processed data, the model performance, its robustness, and its explicability in line with the 2021 Artificial Intelligence (AI) Act~\citep{floridi2021european}. This is a major issue because the acceptance of AI will be possible only if it is trustworthy. On this line, \citet{swischuk2019projection} examined the effectiveness of several metamodeling approaches for predicting high-dimensional flow outputs. They noted the need for interpretation of reduced-order model predictions, but also the choice of the reduced-order model approach depending on available data. The POD/GPR reduced-order model approach we propose in this work is rooted in this need for interpretability and physical understanding of the reduced-order model predictions.   

In this study, the objective of our POD/GPR reduced-order model is to emulate mean (time-averaged) tracer concentration field statistics obtained with LES
for a two-dimensional boundary-layer flow interacting with a surface-mounted obstacle. We consider this simplified test case compared to actual urban canopy to have the capacity to generate a very large LES database and to analyze the reduced-order model accuracy with respect to uncertain inflow boundary conditions and emission source location, while remaining near to validation experiments that have been well-documented through field trials and physical modeling~\citep{peterka1985wind,martinuzzi1993flow}. 

The location of the source emission is highly variable on both sides of the obstacle and at different heights, making the input/output relationship more complex to learn. To overcome this issue, we combine POD and GPR to link the physical space (in which the LES statistics evolve) and the GPR hyperparameters (related to the uncertain input parameters), with the objective to better pose the GPR optimization process that is a key aspect to obtain satisfying prediction performance. In particular, we propose using POD information to fit GPR prior distributions, and thereby accelerate the offline training stage of the POD/GPR reduced-order model. It is worth noting that we mainly focus on mean LES tracer concentration statistics to provide a proof of concept, but an extension to velocity-concentration cross-quantity (i.e.~turbulent scalar flux) is also presented to investigate the capacity of the reduced-order model to emulate other coupled LES statistics. We also identify the minimum learning database required to obtain acceptable reduced-order model performance to prepare for more realistic atmospheric dispersion applications.

The outline of this paper is as follows. Section~\ref{sec:simu} presents the selected test case and the numerical configuration to generate the LES database. In Sect.~\ref{sec:method}, we present our POD/GPR reduced-order modeling approach. Section~\ref{sec:res} analyses in detail the performance of our reduced-order model with focus on POD truncation, prior information for GPR hyperparameter optimization, and sensitivity to training data.

\section{Large-eddy simulation of turbulent flow over a surface-mounted obstacle}
\label{sec:simu}

This section presents the two-dimensional test case and the numerical configuration considered in this study, with a focus on the modeling of inflow boundary conditions and parametric uncertainties. We run an ensemble of simulations with multiple parameter entries to collect LES snapshots from which a reduced-order model can be learned. Despite the fact that the solver we use does not resolve all flow scales (specificity of LES solvers), we will refer to it as a full-order model. 

\subsection{Numerical solver}

The LES database is obtained with the AVBP\footnote{AVBP documentation, see http://www.cerfacs.fr/avbp7x/} LES code developed by CERFACS~\citep{Schonfeld-1999}. AVBP solves the filtered compressible Navier-Stokes equations on unstructured mesh. Additional advection-diffusion equations are solved for passive scalar dispersion. It is widely used to predict non-reactive and reacting unsteady flows in simple or complex geometry, and is applicable to pollutant formation and atmospheric dispersion~\citep{Paoli-2020}. 

In the low-Mach context of atmospheric dispersion, an artificial compressibility approach~\citep{ramshaw1985pressure} is used to increase the time-step resulting from the Courant-Friedrichs-Lewy (CFL) condition. The numerical discretization is based on an explicit, centered scheme from the continuous Taylor-Galerkin family called TTG4A, which is third-order in space and fourth-order in time~\citep{colin2000ttg}. The contribution of subgrid turbulence to the resolved flow variables is modeled with the Smagorinsky closure~\citep{smagorinsky1963}. The subgrid scalar flux is modeled with the Boussinesq hypothesis with a subgrid turbulent Schmidt value equal to 0.7.

\subsection{Case description and boundary conditions}
\label{sec:simu_case}

\begin{figure}[hbt]
\centering
\includegraphics[width=1\textwidth]{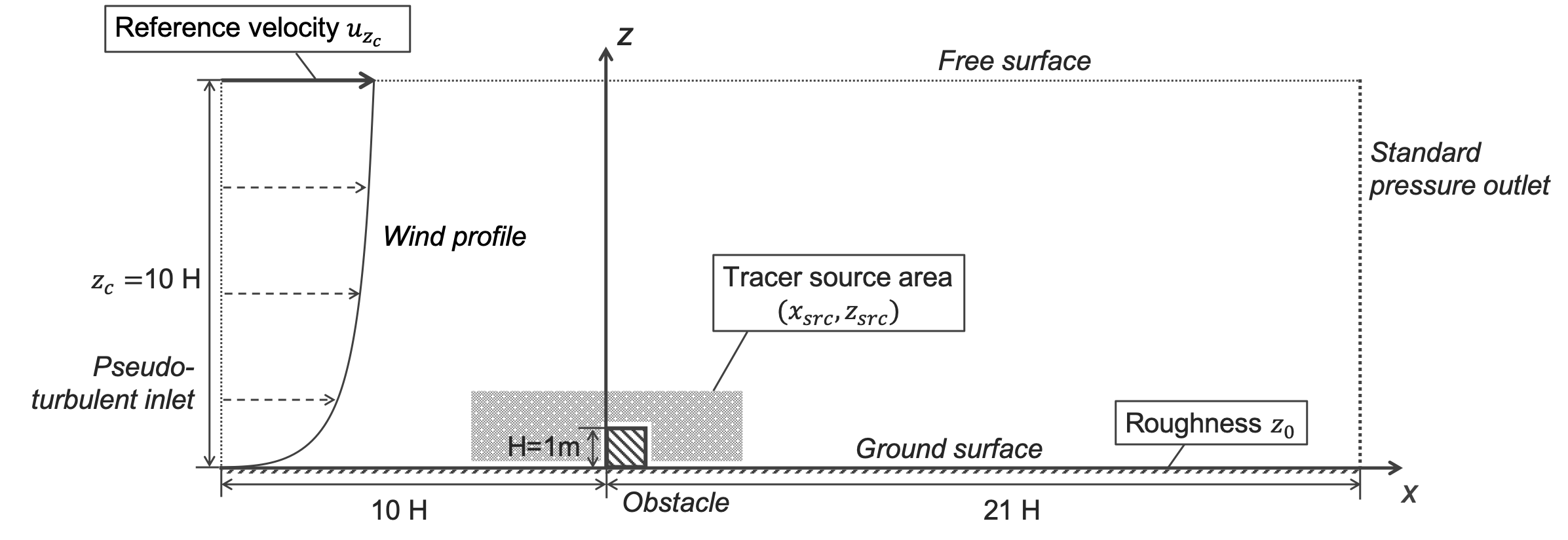}
\caption{Sketch of the test case modeling a turbulent boundary-layer flow interacting with a surface-mounted square obstacle (crosshatched area). Text boxes indicate the uncertain parameters. The gray area indicates the area for the tracer emission source.}
\label{fig:scheme2D}
\end{figure}

This selected test case corresponds to an isolated square-shaped obstacle that is mounted on a rough ground surface~\citep{vincont2000,gamel2015caracterisation}. As depicted in Fig.~\ref{fig:scheme2D}, the obstacle interacts with a developed neutral turbulent boundary-layer flow from the left boundary. 

The height of the obstacle is $H = 1$~m. The two-dimensional computational domain is 31-m long ($x$-axis) by 10-m high ($z$-axis). The domain height is ten times the obstacle height following guidelines for urban flow simulations~\citep{franke2011cost}. It is discretized with a uniform triangular mesh comprising 240,000 elements with a characteristic size $\Delta x = H/10$. 

The left side of the domain (at $x = -10$~m) represents a turbulent inlet boundary where unsteady wind conditions are injected as detailed in Section~\ref{sec:simu_ic_bc}. The right side (at $x = 21$~m) and the upper side (at $z = 10$~m) correspond to an outlet with imposed pressure. The bottom part, including the ground (at $z = 0$~m) and the obstacle surfaces (the obstacle is centered in $(x,z)=[0.5,0.5]$~m$\times$m), corresponds to a rough surface modeled with a law of the wall based on a roughness length $z_0$~[m]. 

The passive gas tracer is released from a point source emission, which can be either located  upstream, above or downstream of the obstacle as seen from the tracer source area in Fig.~\ref{fig:scheme2D}. The release is continuous and constant throughout the simulation, and the emission source fits a Gaussian shape in space with a standard deviation of 0.1~m around the center of release. 

\subsection{Inflow boundary condition modeling}
\label{sec:simu_ic_bc}

\subsubsection{Mean inlet wind profile}
\label{sec:simu_ic_bc_mean}

The inlet wind condition imposes a mean vertical profile $\overline{u}_{inlet}$, based on the Monin-Obukhov similarity theory (MOST) in neutral conditions:
\begin{equation}
\begin{aligned}
\label{eq:logprofile}
\begin{cases}
    \displaystyle\overline{u}_{inlet}(z) = \frac{u_\tau}{\kappa}\,\log\left(1 + \frac{z}{z_0}\right), \\\noalign{\vskip5pt}
    \displaystyle\overline{u}_{inlet}(z = z_c) = u_{z_c},
\end{cases}
\end{aligned}
\end{equation}
where $u_\tau$~[m\,s$^{-1}$] is the friction velocity, $\kappa = 0.41$ is the dimensionless von Kármán constant, $z_0$~[m] is the aerodynamic roughness length, and $z$~[m] is the vertical axis in the domain. In this study, the velocity $u_{z_c}$ enforced at height $z_c = 10$~m and the characteristic surface roughness length $z_0$ are used as input parameters in order to mimic operational conditions, where velocity measurements are typically obtained at some arbitrary reference height~\citep{sousa2019}. From $u_{z_c}$ and $z_0$, Eq.~\ref{eq:logprofile} can be inverted to obtain the corresponding friction velocity $u_{\tau}$: 
\begin{equation}
\displaystyle u_\tau = \frac{\kappa}{\log\left(1+\frac{z_c}{z_0}\right)}u_{z_c}.
\end{equation}
Consistently with the inlet condition, the same surface roughness length $z_0$ is considered assuming a fully developed inlet flow over a rough terrain.

\subsubsection{Inlet wind fluctuations}
\label{sec:simu_ic_bc_turb}

In addition to the mean inlet wind profile from MOST theory, wind fluctuations are superimposed on the mean profile to obtain a turbulent inlet boundary condition that mimics boundary-layer turbulence. The synthetic fluctuations are generated with the Kraichnan method implemented in the AVBP solver~\citep{daviller2019injection}, and follow the Passot-Pouquet turbulence spectrum~\citep{passot1987numerical}. The target turbulent kinetic energy $\mathcal{K}$ at the inlet is estimated as $\mathcal{K}= u_{\tau}^2/C_{\mu}$ with $C_{\mu} = 0.09$~\citep{richards1993}.

\subsubsection{Choice of quantities of interest}

\begin{figure}[!htb]
\centering
\includegraphics[width=0.9\textwidth]{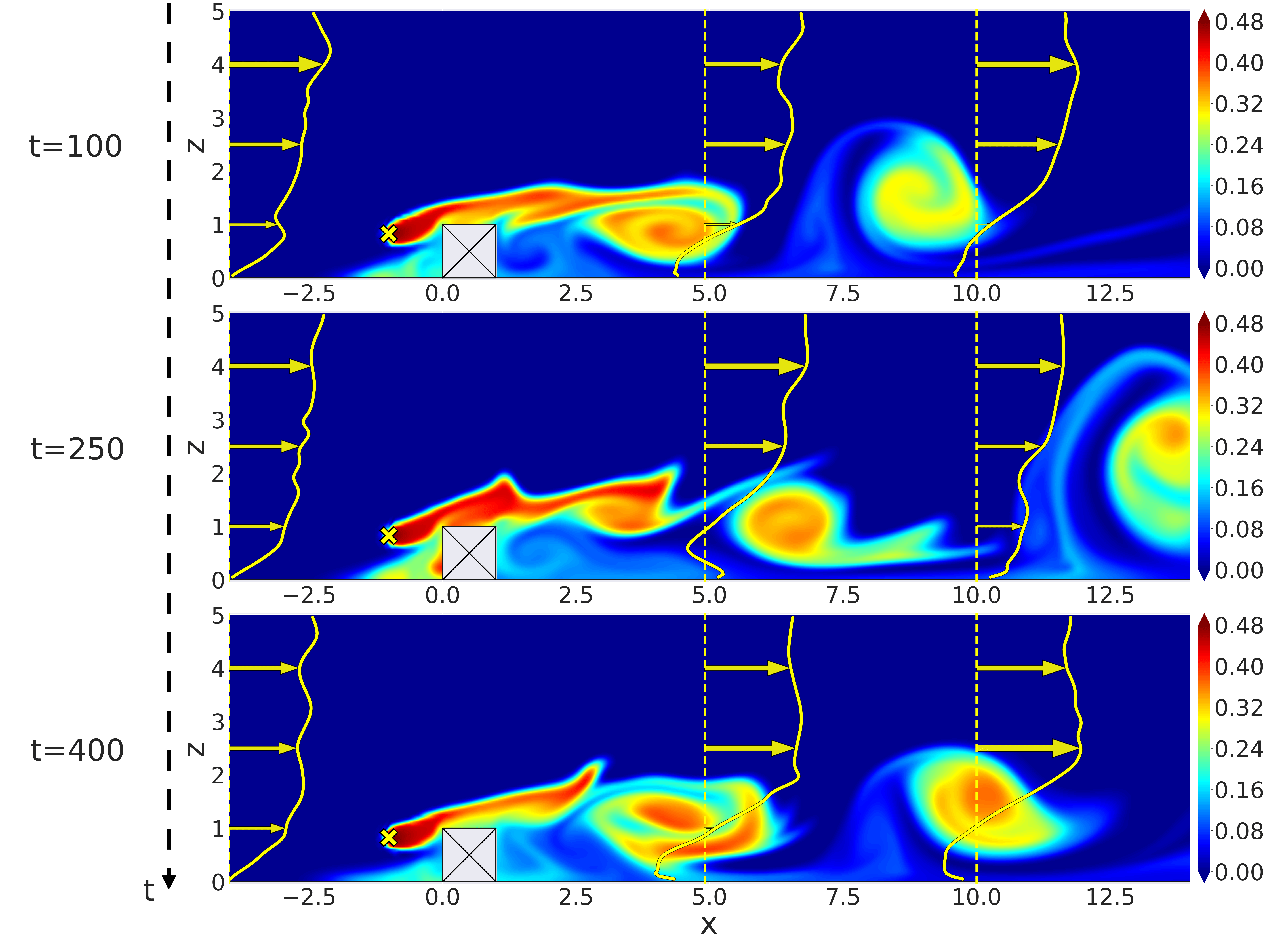}
\caption{Instantaneous normalized tracer concentration fields at instantaneous times $t = 100$, 250 and 400~s from the nominal LES governed by streamwise velocity $u_{z_c}=5.78$~m\,s$^{-1}$ at domain's height $z_c = 10$~m, of ground roughness $z_0 = 2.79\times 10^{-2}$~m, source position $x_{src} = -1.01$~m and source height $z_{src} = 0.830$~m. Three vertical profiles of streamwise velocities at $x =-4$, 5 and 10~m are superimposed on the fields.}
\label{fig:inst_nominal}
\end{figure}

LES is used here to simulate unsteady flow and tracer features, and to obtain reference statistics (temporal mean) of the flow velocity and tracer concentration fields across the domain. Figure~\ref{fig:inst_nominal} shows several instantaneous tracer concentration fields for a given snapshot of the LES dataset. This snapshot can be considered as the nominal solution since it is representative of the average atmospheric conditions considered in this study (Sect.~\ref{sec:simu_uq_param}). The emission source, located upstream and at the height of the square, illustrates the complexity of the dispersion process, which is driven by a combination of the quasi-periodic vortex shedding induced by the flow/obstacle interactions and the inflow turbulence.

\begin{figure}[!htb]
\centering
\begin{overpic}[width=0.8\textwidth]{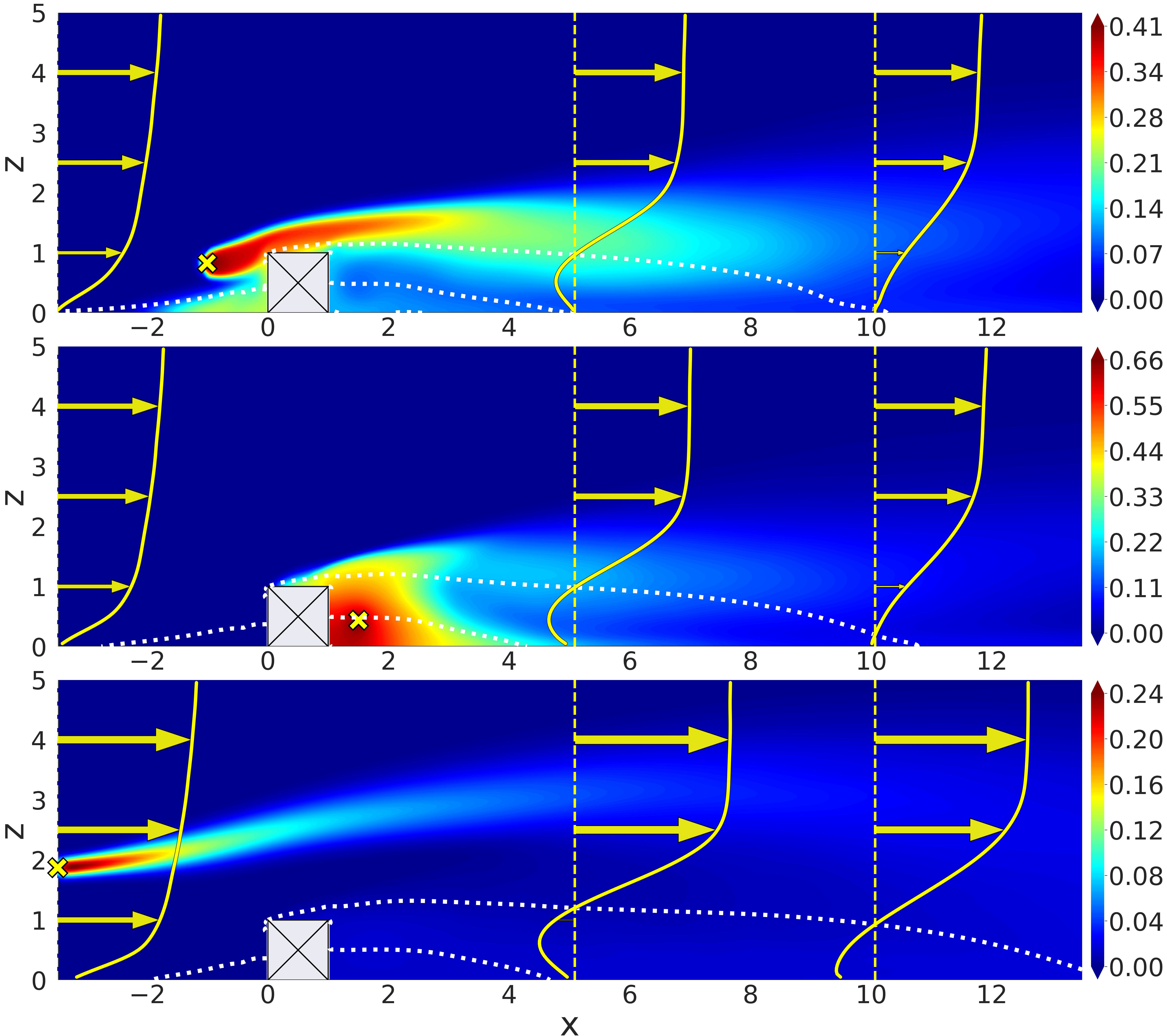}
 \put (-4,80) {\large(a)}
 \put (-4,52) {\large(b)}
 \put (-4,25) {\large(c)}
\end{overpic}
\caption{Mean (time-averaged) normalized tracer concentration field with superimposed time-averaged streamwise velocity vertical profiles at abscissa~$x = -4$, 5 and 10~m (vertical solid lines) and zero-velocity magnitude contour lines (dotted lines) for three LES snapshots of the LES test database: (a)~the nominal snapshot (associated with Fig.~\ref{fig:inst_nominal}); (b)~the snapshot where the streamwise velocity is $U_{z_c}=5.79$~m\,s$^{-1}$, the ground roughness is $z_0=7.89\times 10^{-3}$~m, and the source position and height are $(x_{src},z_{src})=(1.5~\text{m},0.44~\text{m})$; and (c)~the snapshot where the streamwise velocity is $U_{z_c}=7.45$~m\,s$^{-1}$, the ground roughness is $z_0=1.3\times 10^{-3}$~m, and the source position and height are $(x_{src},z_{src})=(-3.49~\text{m},1.86~\text{m})$.}
\label{fig:mean_nominal}
\end{figure}

The main quantities of interest targeted in this study are the time-averaged tracer concentration coefficient spatial fields obtained with LES (an extension to the turbulent scalar flux is also presented in Sect.~\ref{sec:res_fluctu} to highlight the generality of our approach). Figure~\ref{fig:mean_nominal} shows examples of normalized tracer concentration and streamwise velocity mean fields for three parametric scenarios: the nominal case (Fig.~\ref{fig:mean_nominal}a), the case of a source emission placed downstream of the obstacle within a recirculation area (Fig.~\ref{fig:mean_nominal}b), and the case of a high source emission where there is no significant influence of the obstacle on the dispersion (Fig.~\ref{fig:mean_nominal}c). The characteristic scales of the tracer concentration structures in the three configurations are very different, making them good examples to illustrate the strengths and limitations of the proposed approach.  

\subsection{Uncertainty modeling}
\label{sec:simu_uq_param}

\subsubsection{Sources of uncertainty}

The objective of this study is to provide a methodology to explore how input parameter uncertainties propagate on the simulated tracer concentration statistics in the context of boundary-layer flow LES. These parameter uncertainties relate to the specification of the inflow boundary conditions (Sect.~\ref{sec:simu_ic_bc}) and of the tracer emission source. We consider here four uncertain parameters: \emph{i)}~the reference velocity magnitude $u_{z_c}$ at domain's height $z_c = 10\,H$; \emph{ii)}~the aerodynamic roughness length $z_0$; \emph{iii)}~the emission source position $x_{src}$; and \emph{iv)}~the emission source height $z_{src}$. The input vector on uncertainty parameters can be expressed as the four-dimensional vector $\boldsymbol{\mu} = (u_{z_c}, z_0, x_{src}, z_{src})^T$.

\subsubsection{Parametric uncertainty}

\paragraph{Distribution over the inlet wind conditions}

The uncertain roughness length $z_0$ and the reference velocity magnitude $u_{z_c}$ impact the mean inlet wind profile, implying that the inlet wind $u(x_{inlet},z)$ becomes uncertain.

The range for the roughness length is set based on small obstacles between fallow ground and low mature agricultural crops~\citep{wieringa1992updating}, leading to $z_0\in[10^{-3},10^{-1}]$~m. Its distribution is log-uniform with probability density function:
\begin{equation}
\label{eq:z0_distribution}
\log(z_0) \sim \mathcal{U}\left(\log(10^{-3}), \log(10^{-1})\right),
\end{equation}
corresponding to a mean value $\mathbb{E}[z_0] \approx 0.021~\mathrm{m}$ and a standard deviation $\sigma(z_0) \approx 0.025~\mathrm{m}$ (the coefficient of variation, i.e.~the ratio of the mean value to the standard deviation, is 1.16). Note that the distribution on $z_0$ is defined as log-uniform so that the marginal distribution $u(x_{inlet},z) \lvert u_\tau$ is close to a uniform distribution.

Streamwise velocity magnitude at domain's height $u_{z_c}$ is supposed to follow a uniform distribution so that the marginal distribution $u(x_{inlet},z) \lvert z_0$ is also uniform:
\begin{equation}
\label{eq:uzc_distribution}
u_{z_c} \sim \mathcal{U}([3,9])~\text{m\,s}^{-1},
\end{equation}
corresponding to a mean value $\mathbb{E}[u_{z_c}] = 6.0~\mathrm{m}\,\mathrm{s}^{-1}$ and a standard deviation $\sigma(u_{z_c}) \approx 1.7~\mathrm{m}\,\mathrm{s}^{-1}$ (the corresponding coefficient of variation is 0.29). The range of variation for $u_{z_c}$ is chosen in agreement with urban air dispersion studies in the literature~\citep{garcia2014quantifying}.

We need to normalize the quantities of interest to train the reduced-order model. For this purpose, we define the reference velocity $u_\tau^{(ref)}$ as the mean value of $u_\tau$ over the ensemble obtained by perturbing both the roughness length $z_0$ and the streamwise velocity magnitude $u_{z_c}$ (using Monte Carlo random sampling and following the statistical distributions in Eqs.~\ref{eq:z0_distribution}--\ref{eq:uzc_distribution}). We normalize the quantities of interest using this reference velocity $u_\tau^{(ref)}$ and a reference length scale (the obstacle's height $H = 1$~m) as:
\begin{equation}
\widetilde{u} = \frac{u}{u_\tau^{(ref)}}, \quad \widetilde{K} = K\left(\frac{u_\tau^{(ref)}\,\mathrm{H}^2}{Q_s}\right),
\end{equation}
where the normalized velocity and tracer concentration are denoted by $\widetilde{u}$ and $\widetilde{K}$, respectively, with the reference velocity $u_\tau^{(ref)} = \mathbb{E}[u_\tau] \approx 3.7\times 10^{-1}$~m\,s$^{-1}$ and the flow rate $Q_s = 1$~m$^{3}$\,s$^{-1}$. In the following, we drop the tilde notation for the sake of simplification.

\begin{figure}[!htb]%
\centering
\includegraphics[width=0.8\textwidth]{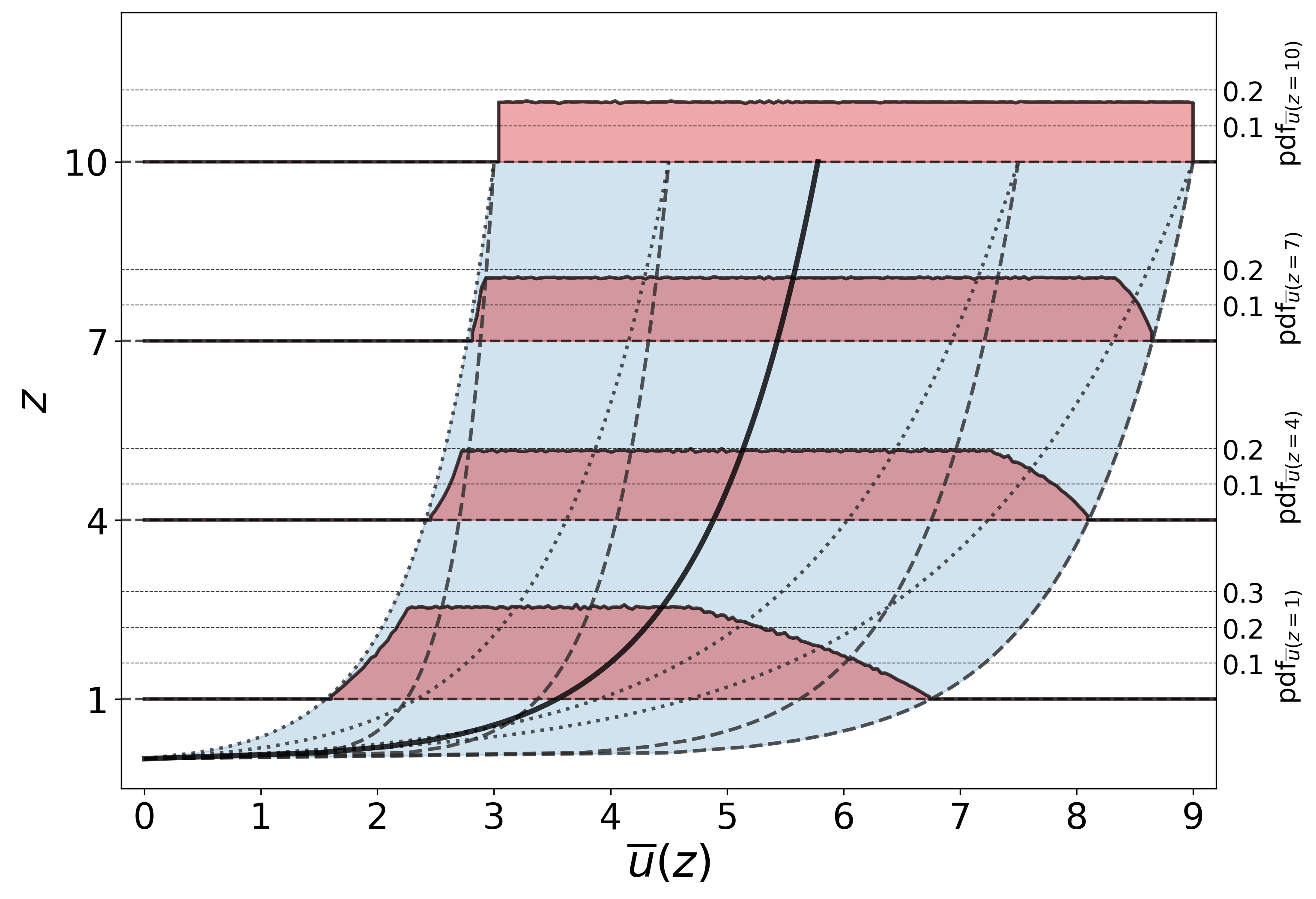}
\caption{Probability distribution of $\overline{\mathbf{u}}$-uncertain space and more specifically, of the mean streamwise velocity inlet profile $\overline{u}_{inlet}(z)$ at $x=-10~\text{m}$. The shaded region denotes the bounded support of non-zero probability. Streamwise velocity probability distributions are plotted at heights $z = 1$, 4, 7, 10~m. Examples of logarithmic profiles are shown for $z_0 = 10^{-3}$~m (dashed lines) and $z_0 = 10^{-1}$~m (dotted lines). The nominal case (Fig.~\ref{fig:mean_nominal}a) profile is also plotted (solid line).}
\label{fig:log_profile}
\end{figure}

The probability distribution on the mean inlet wind profile $\overline{u}(x_{inlet},z)$ depends on the marginal distributions chosen on the two uncertain parameters $z_0$ and $u_{z_c}$. Figure~\ref{fig:log_profile} shows how this distribution varies with respect to the vertical axis $z$. At domain's height $z = 10$~m, it follows a uniform distribution corresponding to the distribution on $u_{z_c}$. But this is no longer the case when getting closer to the ground surface (i.e. when $z < 10$~m), the probability distribution support decreasing as $z$ decreases. Several mean inlet wind profiles based on varying entries for $u_{z_c}$ and $z_0$ are also plotted in Fig.~\ref{fig:log_profile} to illustrate the variety of profiles in the LES ensemble. 

\paragraph{Distribution over the tracer emission source characteristics}

Tracer uncertainty relates to the horizontal and vertical coordinates of the source location $(x_{src}, z_{src})$, also referred to as position and height, respectively. In this study, tracer emission can occur upstream, downstream or above the obstacle. The range of variation is chosen so as to cover a broad panel of existing experimental studies~\citep{mavroidis2007atmospheric,gamel2015caracterisation,du2020novel}. The marginal distributions associated with the source position and height are set uniform as:
\begin{equation}
\label{eq:xzsrc_distribution}
x_{src} \sim \mathcal{U}([-3.5,3.5])~\text{m}, \quad z_{src} \sim \mathcal{U}([0.2,2.0])~\text{m}.
\end{equation}
The range of variation for the source height is slightly away from the ground surface to avoid boundary issues. The area $[0,1.2]\times[-0.2,1.2]$ is removed from the range of variation to avoid having a source inside the obstacle.

\subsection{Full-order model database}
\label{sec:les_fullorder_form}

\subsubsection{Dataset acquisition}

In this study, we consider different sources of uncertainty in the LES full-order model: \emph{i)}~uncertainties associated with the large-scale atmospheric flow conditions $\boldsymbol{\mu}_{atm}$ (affecting the inflow and surface boundary conditions), and \emph{ii)}~uncertainties with the tracer emission source characteristics $\boldsymbol{\mu}_{tr}$. These uncertainties are described by uncertain scalar parameters that are considered as inputs to the LES problem (Sect.~\ref{sec:simu_uq_param}), and that form the input vector parameter: 
\begin{equation}
\boldsymbol{\mu}=(\boldsymbol{\mu}_{atm},\boldsymbol{\mu}_{tr})^T = (u_{z_c}, z_0, x_{src}, z_{src})^T \in \mathbb{R}^4.
\end{equation}
The input parameters impact the simulated flow response and drive the quantities of interest $\mathbf{K}_\mathrm{les} = \{{K}_{1},\hdots,{K}_{N_{h}}\}^T \in \mathbb{R}^{N_{h}}$, namely means of tracer concentrations predicted at the $N_h$ grid elements of the discretized computational domain (the nominal snapshot example is given in Fig.~\ref{fig:mean_nominal}a).

The LES full-order model provides accurate prediction of the quantities of interest but at a significant computational cost, which prevents real-time outputs and requires extensive computational resources (the average cost of one simulation is about 700~CPU hours). These issues motivate the use of a reduced-order model, which can result in significant speedups to predict the quantities of interest for any new value of the input vector $\boldsymbol{\mu}$. Therefore, the objective of the reduced-order model is to approximate the LES model response based on a collection of $N$ snapshots (or samples) $\lbrace \mathbf{K}_\mathrm{les}^{(1)}, \cdots, \mathbf{K}_\mathrm{les}^{(N)} \rbrace$ corresponding to a collection of $N$ input sets $\lbrace \boldsymbol{\mu}^{(1)}, \cdots, \boldsymbol{\mu}^{(N)} \rbrace$ with $N \ll N_h$. One challenge for training and validating the reduced-order model is that the number of snapshots $N$ remains limited in the context of LES.

\subsubsection{Experimental sampling design}
\label{sec:sampling_strategy}

In this study, we build a dataset of 750~LES in order to train and carefully test our reduced-order model (this dataset can be considered as large from a LES perspective but still remains limited from a statistical learning perspective). Each snapshot of the dataset corresponds to different realizations of the input vector $\boldsymbol{\mu} = (u_{z_c}, z_0, x_{src}, z_{src})^T$. The samples are obtained using Halton's low-discrepancy sequence, since the number of input parameters remains small and we need to have a good coverage of the uncertain space for a limited number of samples. Note that we restrict the area of interest on the $x$-axis to the interval $[-3.5,13.5]~\text{m}$, containing 99.9\% of the overall ensemble variance (this area still encloses the space of uncertainty on the source location position $x_{src}$ and height $z_{src}$). Moreover, all LES were conducted on the same numerical setting (same grid, convection scheme, LES model, etc.). Since the LES setup is robust, the non-intrusive procedure is well decoupled from the LES model to only handle the parametric variability.

The physical time of simulation necessary to achieve converged LES statistics is one challenge with the LES database generation to avoid introducing noise during the training stage. For each snapshot, the initial flow field is specified from the inlet boundary condition, and a spin-up is then necessary to establish turbulence throughout the computational domain and to obtain LES field statistics that are consistent with the presence of the obstacle. Characteristic time scales of the flow are extracted for each snapshot (the flow periodicity is based on the Strouhal number). The time averaging process of the LES outputs is performed on forty characteristic time periods to limit the noise associated with LES solver convergence.

\section{Non-intrusive reduced-order modeling}
\label{sec:method}

A direct numerical approximation of the full-order model is not affordable in the many-query context of parameterized LES. This section describes the non-intrusive reduced-order modeling strategy that we deploy for learning the LES model response on the space of the input parameters $\boldsymbol{\mu}$. This is carried out in the statistical learning framework introduced by~\citet{guo2018reduced}, where the learning stage (offline phase) corresponds to the construction (training) and evaluation (validation) of the reduced-order model. Later, multi-query evaluations of the reduced-order model (inline phase) allow predictions of additional scenarios and are thereby independent of the full-order model dimension. We focus next on the learning stage.

\subsection{Reduced basis}

The purpose of reduced-basis approaches is to provide statistically efficient approximate solutions of the full-order model, decreasing the computational burden while minimizing the loss of accuracy. The model is built from a set of full-order snapshots and results in a reduced-basis space of representation of reduced-basis functions (or modes) $\mathcal{V}_\mathrm{rb} = \mathrm{Span}(\{\boldsymbol{\psi}_l\}_{l=1,\hdots,L})\subset \mathbb{R}^{N_{h}}$. This space is assumed to be of low dimension compared to the number of grid elements (i.e. $L \ll N_{h}$). Reduced-basis solutions are usually computed from linear combinations of the modes:
\begin{equation}
\begin{aligned}
    \mathbf{G}_\mathrm{rb}\colon \mathcal{P} & \longrightarrow \mathcal{V}_\mathrm{rb}\\
    \boldsymbol{\mu} & \longmapsto \mathbf{K}_\mathrm{rb} = \displaystyle\sum_{l = 1}^L\,k_l(\boldsymbol{\mu})\,\boldsymbol{\psi}_l(x,z) \, , \label{eq:rom}
\end{aligned}
\end{equation}
where $k_l(\boldsymbol{\mu})\in\mathbb{R}$ denotes the $l$th reduced coefficient. 

This approach provides a procedure to split the parametric dependency in $\boldsymbol{\mu}$ from the spatial dimension $(x,z)$ since local dissimilarity is now carried by the modes. In a first step, the modes are obtained from POD. This procedure transforms LES data (i.e.~projects the LES snapshots in a reduced space spanned by a set of parameter-independent functions $\lbrace \boldsymbol{\psi}_l \rbrace_{l = 1, \cdots, L}$), and thereby returns discrete reduced-basis coefficients from the original output fields (Sect.~\ref{sec:build_pod}). 
In a second step, regression metamodels based on GPR are trained to map the uncertain parameters $\boldsymbol{\mu}$ onto the reduced coefficients $\lbrace k_l \rbrace_{l = 1, \cdots, L}$ (Sect.~\ref{sec:meth_gpr}). The resulting reduced-order model (Eq.~\ref{eq:rom}) can finally be used to estimate new quantities of interest $\mathbf{K}_\mathrm{rb}$ at unexplored parameter values $\boldsymbol{\mu}^*$ (i.e. at parameter values that are not included in the LES training database). 

The difficulties in building the reduced-order model are three-fold: \emph{i)}~the quantities of interest simulated using LES are of very large dimension ($N_{h}$ is on the order of $10^5$ in this study); \emph{ii)}~the number of snapshots is limited due to the computational cost of a single LES (i.e.~$N \ll N_{h}$); and \emph{iii)}~the mapping between the quantities of interest $\mathbf{K}$ and the input parameters $\boldsymbol{\mu}$ may be subject to nonlinearity due to changes in the emission source location and associated changes in the shape and position of the tracer concentration wakes.

\subsection{Proper orthogonal decomposition}
\label{sec:build_pod}

\subsubsection{Optimal reduced basis and dimension reduction}
\label{sec:build_pod_def}

In practice, POD~\citep{sirovich1987,berkooz1993} computes an estimate of the optimal solution $\mathcal{V}_\mathrm{rb}^*$ from a collection of LES snapshots gathered in the snapshot matrix: 
\begin{equation}
\mathbf{S} = \left[\mathbf{K}_\mathrm{les}^{(1)} \mid \hdots \mid \mathbf{K}_\mathrm{les}^{(N)}\right] \in \mathbb{R}^{N_h \times N}.
\end{equation}
POD seeks the optimal reduced basis of rank $L$ that stands as the optimal orthogonal projection manifold with respect to the Frobenius norm:
\begin{align}
    \mathcal{V}_\mathrm{rb}^* = \mathrm{argmin}_{\substack{\noalign{\vskip1pt}
        \mathbf{S}_\mathrm{rb}\;=\; p(\mathbf{S};\mathcal{V}_\mathrm{rb}),\\\noalign{\vskip1pt}
        \mathrm{rank}(\mathcal{V}_\mathrm{rb})\;=\;L
    }}\|\mathbf{S}-\mathbf{S}_\mathrm{rb}\|_{\mathrm{F}},
\end{align}
with $p(\cdot;\mathcal{V}_\mathrm{rb})$ the projection from $\mathbb{R}^{N_{h}}$ to $\mathcal{V}_\mathrm{rb}$.
The idea behind POD is to express an orthonormal basis maximizing the variance of the projected field ensemble. This problem may be solved by diagonalizing the covariance matrix:
\begin{equation}
\begin{aligned}
\label{eq:pod_diagonalization}
    & \mathrm{Cov}(\mathbf{K}_\mathrm{les},\mathbf{K}_\mathrm{les}) = \mathbf{S}\,\mathbf{S}^T = \Psi\,\Sigma^2\,\Psi^T,\\
    & \mathrm{with}~\begin{cases}
        \mathrm{Cov}(\mathbf{K}_\mathrm{les},\mathbf{K}_\mathrm{les}) \in \mathbb{R}^{N_{h}\times N_{h}}
        \\\noalign{\vskip5pt}
        \Psi=[\boldsymbol{\psi}_1~\lvert~\hdots~\rvert ~\boldsymbol{\psi}_{N_{h}}]\in\mathbb{R}^{N_{h}\times N_{h}}~\text{an orthonormal matrix} \\\noalign{\vskip5pt}
        \Sigma = \mathrm{diag}(\sigma_1,\hdots,\sigma_{N_{h}})\in\mathbb{R}_+^{N_{h}\times N_{h}},~\sigma_1\ge \cdots \ge\sigma_{N_{h}} > 0.
    \end{cases}
\end{aligned}
\end{equation}
Orthonormal vectors of $\Psi$, called the reduced-basis modes, carry some fraction of the ensemble variance quantified by the related eigenvalues in $\Sigma$ denoted by $\lbrace\sigma_l\rbrace_{l = 1, \cdots, N_h}$ and satisfying $\mathbf{S}\,\mathbf{S}^T\,\boldsymbol{\psi}_l = \sigma_l\,\boldsymbol{\psi}_l$. 

Since the number of mesh elements $N_h$ is very large, it becomes advantageous to keep the $L$ first modes (among the $N_h$ modes) that preserve the maximum variance of the original ensemble. The resulting truncated matrices are denoted by $\widetilde{\Psi} =[\boldsymbol{\psi}_1~\lvert~\hdots~\rvert ~\boldsymbol{\psi}_{L}]\in\mathbb{R}^{N_{h}\times L}$ and $\widetilde{\Sigma} = \mathrm{diag}(\sigma_1,\hdots,\sigma_{L}) \in\mathbb{R}^{L\times L}$. We investigate here the number of modes $L$ that is necessary to maintain in the reduced basis to obtain an accurate reduced-order model over the input parameter space and in the different areas of interest of the computational domain (peak tracer concentration near the emission source upstream of the obstacle, tracer accumulation upstream and downstream of the obstacle, tracer dispersion downstream of the obstacle).

\subsubsection{Mode interpretation}
\label{sec:build_pod_correl}

POD acts as a change of basis and the meaning of each mode can be interpreted by a correlation analysis between the modes and the original features. In practical terms, the correlation between the $l$th POD mode $\boldsymbol{\psi}_l$ and a given snapshot in the initial grid space $\mathbf{K}_\mathrm{les}$ (varying between -1 and 1 by definition) may be stated as the following matrix:
\begin{equation}
\label{eq:pod_correlation}
\mathrm{Corr}(\mathbf{K}_\mathrm{les},\boldsymbol{\psi}_l) = \left[\sqrt{\frac{\sigma_l}{\hat{\mathbb{V}}(\mathbf{K}_{\mathrm{les},i})}} \psi_{l,j}\right]_{ij},
\end{equation}
where the indices $i$ and $j$ correspond to a given element of the matrix $\mathrm{Corr}(\mathbf{K}_\mathrm{les},\boldsymbol{\psi}_l) \in \mathbb{R}^{N_h \times N_h}$, where $\psi_{l,j}$
corresponds to the $j$th element of $\boldsymbol{\psi}_l \in \mathbb{R}^{N_h}$, the $l$th function in $\widetilde{\Psi}$, and where $\hat{\mathbb{V}}(\mathbf{K}_{\mathrm{les},i})$ represents the variance unbiased estimation over the snapshots for the $i$th grid element. This variance is estimated as:
\begin{equation}
\hat{\mathbb{V}}(\mathbf{K}_{\mathrm{les},i}) = \frac{1}{N-1}\,\displaystyle\sum_{n=1}^N\,\left(K_i^{(n)} - \hat{\mathbb{E}}(\mathbf{K}_{\mathrm{les},i})\right)^2,~\hat{\mathbb{E}}(\mathbf{K}_{\mathrm{les},i}) = \frac{1}{N}\sum_{n=1}^{N} K_{i}^{(n)},
\label{eq:var_mean_ensemble}
\end{equation}
following notations introduced in Sect.~\ref{sec:les_fullorder_form}.

\subsubsection{Reduced-coefficient dataset}
\label{sec:build_pod_dataset}

The quantity of interest vector field can be projected upon the POD space. The projection can be expressed as:
\begin{equation}
\begin{aligned}
\label{eq:whitening}
    \mathcal{W}\colon \mathbb{R}^{N_{h}} & \longrightarrow \mathbb{R}^{L}\\
    \mathbf{K}_{\mathrm{les}} &\longmapsto \boldsymbol{k} = \widetilde{\Sigma}^{-1/2}\,\widetilde{\Psi}^T\,\mathcal{T}(\mathbf{K}_{\mathrm{les}}),
\end{aligned}
\end{equation}
where $\widetilde{\Sigma}^{-1/2} = \mathrm{diag}(1/\sqrt{\sigma_1},\hdots,1/\sqrt{\sigma_{N_{h}}}) \in\mathbb{R}^{L\times L}$ and $\widetilde{\Psi}\in\mathbb{R}^{N_h\times L}$ are the matrices restricted to the $L$ first modes (Sect.~\ref{sec:build_pod_def}), $\boldsymbol{k} = [k_1, \cdots, k_L] \in\mathbb{R}^L$ is the vector of POD reduced coefficients (Eq.~\ref{eq:rom}), and $\mathcal{T}(\cdot)$ is the operator applying an affine transformation (centering and normalization) to the snapshot data:
\begin{equation}
\begin{aligned}
    \mathcal{T}\colon \mathbb{R}^{N_{h}} & \longrightarrow \mathbb{R}^{N_{h}}\\
    \mathbf{K}_{\mathrm{les}} & \longmapsto \frac{1}{\sqrt{N-1}}(\mathbf{K}_{\mathrm{les}} - \hat{\mathbb{E}}(\mathbf{K}_{\mathrm{les}})),
\end{aligned}
\end{equation}
where $\hat{\mathbb{E}}(\mathbf{K}_{\mathrm{les}}) = \left[\hat{\mathbb{E}}(\mathbf{K}_{\mathrm{les},1}),~\hdots~,~\hat{\mathbb{E}}(\mathbf{K}_{\mathrm{les},N_h})\right]\in\mathbb{R}^{N_{h}}$ is the mean of the quantity of interest over the snapshots for each grid element (Eq.~\ref{eq:var_mean_ensemble}).

The linearity in the operator $\mathcal{T}$ makes it simple to express the inverse reconstruction operator from which the original quantity of interest field may be recovered from the compressed information carried by the POD reduced coefficients:
\begin{equation}
    \mathbf{K}_\mathrm{rb}(\boldsymbol{\mu}) = \mathcal{T}^{-1}\left(\displaystyle\sum_{l = 1}^L~\sqrt{\sigma_l}\,k_l\,\boldsymbol{\psi}_l\right).
\label{eq:pod_inverse}
\end{equation}

One key aspect of our POD implementation is to center and standardize the reduced coefficients obtained in Eq.~\ref{eq:whitening}. The standardization step is usually referred to as whitening (e.g.~\citeauthor{kessy2018optimal}, \citeyear{kessy2018optimal}) and guarantees the following properties for $\boldsymbol{k}$:
\begin{align}
\begin{cases}
\mathbb{E}[k_i] = 0 \quad \forall~i=1,\hdots,L \, , \\ \noalign{\vskip5pt}
\mathbb{E}[k_i\,k_j] 
= \begin{cases}
    1\,\mathrm{if}~i = j, \\
    0\,\mathrm{otherwise}.
\end{cases}\quad \forall~i,j = 1,\hdots, L.
\end{cases}
\label{eq:k_bias_correl}
\end{align}
The transformation in Eq.~\ref{eq:whitening} may now be applied to the original dataset (expressed in the grid space) to create a new dataset of POD reduced coefficients $\boldsymbol{k}$ for varying input parameters $\boldsymbol{\mu}$. The compressed information is suitable for modeling (through a metamodel) the relationship between the mean tracer concentration field and the uncertain input parameters. This is detailed in next section.

\subsection{Gaussian process regression}
\label{sec:meth_gpr}

In this work, rather than training a cumbersome metamodeling procedure involving the high-dimensional LES output fields, we train metamodels to predict the reduced coefficients $\boldsymbol{k} = [k_1, \cdots, k_L]$ from the input parameters $\boldsymbol{\mu}$ through GPR~\citep{williams2006gaussian}.
It is worth mentioning that preliminary tests (not presented here) seemed to show a better performance of GPR models compared to other metamodels based on polynomial chaos expansion and decision trees for instance. This should be confirmed by a more detailed analysis. However, such an intercomparison is beyond the scope of this study, which focuses more on the link between POD and GPR.

\subsubsection{Metamodel formulation}
\label{sec:gp_definition}

\paragraph{Learning task}

POD decomposes ensemble variance into hierarchical information carried by the $L$ reduced coefficients. The first modes carry the large energetic structures of the data, whereas the higher modes focus more and more on local effects. Moreover, the parametric variability induces strong changes in the modal decomposition  distribution. It is for instance obvious that a change in the geometric position of the tracer emission is going to drastically affect the amplitude and/or the signs of the POD coefficients. As a result, the associated response surfaces  $\boldsymbol{k} \equiv  \boldsymbol{k}(\boldsymbol{\mu})$ tend to reflect the characteristic scales of the tracer concentration patterns.

We aim to construct GPR models that take into consideration this prior knowledge (e.g. typical length-scales) in order to be efficient at accurately and robustly describe the POD reduced coefficients. Therefore, one key aspect of the metamodeling stage deals with the choice of the kernel (Sect.~\ref{sec:gpr_kernel_spec}) and the hyperparameter optimization (Sect.~\ref{sec:gp_optim_procedures}) to adapt the Gaussian processes to the variety of length-scales across the POD modes. 

\begin{figure}[hbt]
\centering
\includegraphics[width=1.0\textwidth]{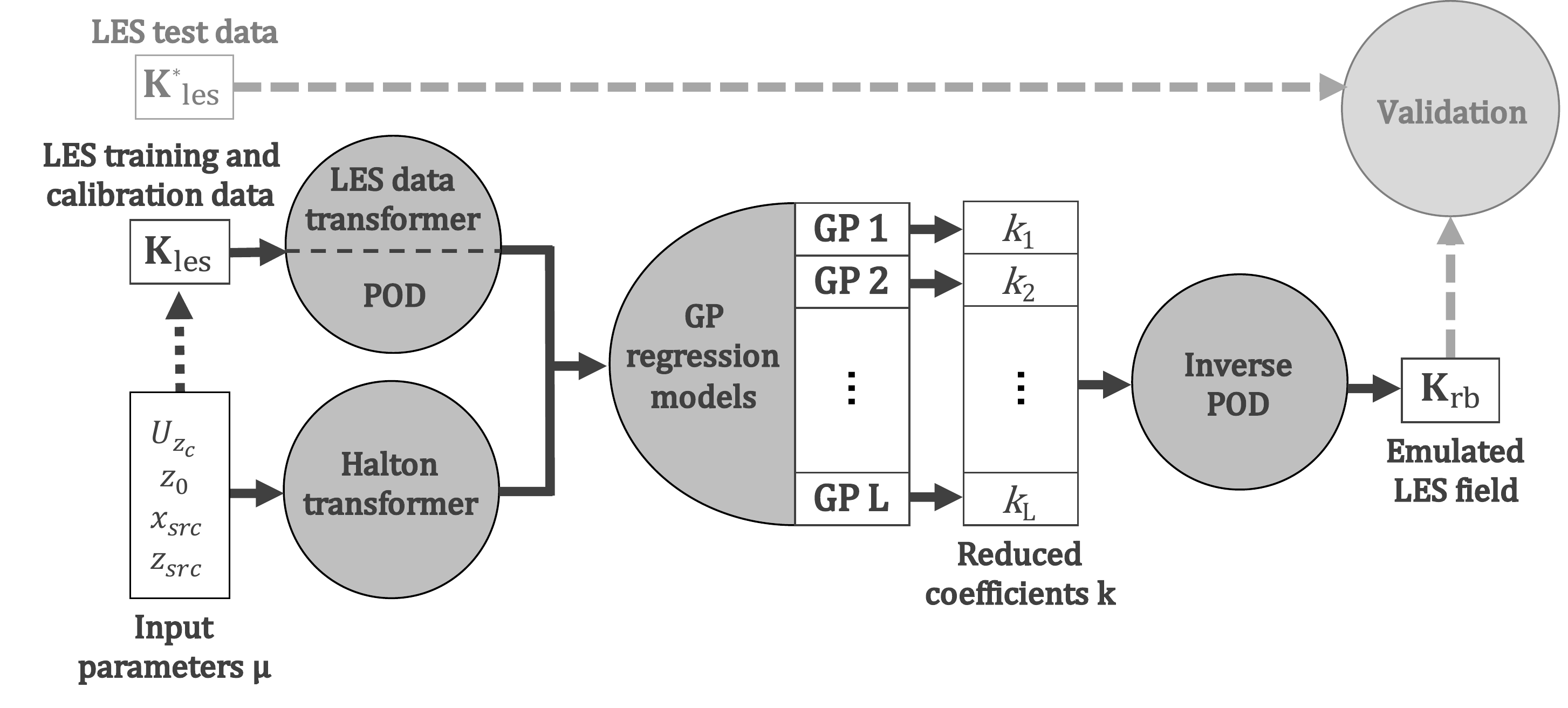}
\caption{Schematic of the reduced-order modeling approach consisting in training $L$ independent GPR models to emulate the $L$ reduced coefficients $[k_1, \cdots, k_L]$ with respect to the input parameters $\boldsymbol{\mu}$, and then to reconstruct the LES field of interest by an inverse POD transformation (the emulated LES field can be compared to the LES test dataset for validation).}
\label{fig:compound_scheme}
\end{figure}

Because the POD reduced coefficients are decorrelated (Eq.~\ref{eq:k_bias_correl}), we design $L$ independent GPR models, i.e.~we learn the relation between each reduced coefficient $k_l$ and the input parameters $\boldsymbol{\mu}$ for $l$ varying from 1 to $L$ (Fig.~\ref{fig:compound_scheme}).

\paragraph{Noisy training data}

We consider the LES data to be somewhat noisy due to time averaging. To account for this noise, the non-interpolating GPR model for the $l$th reduced coefficient $k_l$ is written as:
\begin{equation}
k_l = f_l(\boldsymbol{\mu}) + \epsilon_l,
\label{eq:eq_noise}
\end{equation}
where $\epsilon_l \sim \mathcal{N}(0,s_l^2)$ is an additive noise term with variance $s_l^2$.

\paragraph{Reduced coefficient prediction}

The GPR framework assumes that the mapping $f_l$ is a Gaussian stochastic process such that:
\begin{equation}
f_l(\boldsymbol{\mu}) \sim \mathcal{GP}\left(m_l(\boldsymbol{\mu}), r_l (\boldsymbol{\mu},\boldsymbol{\mu}^*)\right) \quad \forall \, (\boldsymbol{\mu},\boldsymbol{\mu}^*)\in\mathcal{P}\times \mathcal{P},
\label{eq:gp_process}
\end{equation}
with $m_l(\boldsymbol{\mu}) =\mathbb{E}[f_l(\boldsymbol{\mu})]$ the mean function of the Gaussian process and 
$r_l(\boldsymbol{\mu},\boldsymbol{\mu}^*) = \mathbb{E}[(f_l(\boldsymbol{\mu})-m_l(\boldsymbol{\mu})) \, (f_l(\boldsymbol{\mu}^*)-m_l(\boldsymbol{\mu}^*))]$ its associated covariance function. 

GPR starts with a prior distribution over the mean and covariance function. In this work, the prior mean is assumed to be 0 because of POD data transformation (Sect.~\ref{sec:build_pod}), and the prior covariance is specified by choosing a kernel (Sect.~\ref{sec:gpr_kernel_spec}). The learning stage consists in updating the mean and covariance by integrating the information from the reduced-coefficient dataset (Sect.~\ref{sec:build_pod_dataset}). 

We note $\mathcal{D}_l = \{(\boldsymbol{\mu}^{(n)},k^{(n)}_l),~n = 1,\hdots,N\}$ the dataset dedicated to the $l$th reduced coefficient. It is worth noting that we use the raw samples of the Halton's sequence over the interval $[0,1]^4$ as the input data to GPR in order to ensure that the samples are adequately distributed over the uncertain space. This is motivated by the fact that the raw Halton's sequence has suitable statistical properties (same parameter range $[0,1]$, approximately identical distribution), while the input parameters are of very different nature. In particular, the surface roughness $z_0$ is log-uniformly distributed and orders of magnitude smaller in comparison to the other input parameters.   

From now on, we make the distinction between the training sample and the test sample: the corresponding $N_\mathrm{train}$ and $N_\mathrm{test}$-collection matrices are denoted by $(\mathcal{U},\, \mathcal{K}_l)$ and $(\mathcal{U}^*, \,\mathcal{K}_l^*)$. Using these notations, the joint distribution between the training dataset and some new test evaluations is expressed with respect to the kernel as:
\begin{align}
    \begin{bmatrix} 
        \mathcal{K}_l \\ 
        \mathcal{K}_l^*
    \end{bmatrix} \sim
    \mathcal{N}\left(0,
        \begin{bmatrix}
            r_l(\mathcal{U},\mathcal{U})+s_l^2\,I & r_l(\mathcal{U},\mathcal{U}^*)\\
            r_l(\mathcal{U}^*,\mathcal{U}) & r_l(\mathcal{U}^*,\mathcal{U}^*)
        \end{bmatrix}\right),    
\label{eq:gp_kernel}
\end{align}
where $s_l$ is the noise variance (Eq.~\ref{eq:eq_noise}) and $I$ stands for the identity matrix. We can derive the inference formula for the test reduced coefficients from the following conditional distribution:
\begin{equation}
\label{eq:posterior_gpr}
    \mathcal{K}_l^*\,\mid\, \mathcal{U},\mathcal{K}_l,\mathcal{U}^* \sim \mathcal{N}\left(m_l^*,\text{cov}(\mathcal{K}_l^*)\right),
\end{equation}
where 
\begin{equation}
\begin{cases}
m_l^* &= r_l(\mathcal{U}^*,\mathcal{U})\,\left[r_l(\mathcal{U},\mathcal{U}) + s_l^2\,I\right]^{-1}\,\mathcal{K}_l \\
\text{cov}(\mathcal{K}_l^*) &= r_l(\mathcal{U}^*,\mathcal{U}^*) - r_l(\mathcal{U}^*,\mathcal{U})\, \left[r_l(\mathcal{U},\mathcal{U}) + s_l^2\,I\right]^{-1} \,r_l(\mathcal{U},\mathcal{U}^*).
\end{cases}
\label{eq:posterior_gpr2}
\end{equation}
All terms in Eq.~\ref{eq:posterior_gpr2} are known. The covariance formulation depends on prior variance $r_l(\mathcal{U}^*,\mathcal{U}^*)$ over the test dataset refined by information from the training dataset. In this formulation, the matrix $[r_l(\mathcal{U},\mathcal{U}) + s_l^2\,I]$ is inverted using a computationally-efficient Cholesky decomposition~\citep{williams2006gaussian}. Therefore, the posterior distribution for the $l$th reduced coefficient can be directly estimated using Eq.~\ref{eq:posterior_gpr}. The GPR estimator is set as the mean posterior, which is a linear combination of kernel distances computed between the test point and all the training data. 

\subsubsection{Anisotropic Matérn kernel}
\label{sec:gpr_kernel_spec}

The choice of the kernel is crucial as it entails specific assumptions on data covariance in the input space $\mathcal{P}$. In this work, we consider the Matérn class of covariance functions, which features interesting stability and smoothness properties (via hyperparameters) for machine learning. 

\paragraph{Matérn kernel definition}

The Matérn kernel leads the underlying process to be stationary since it is expressed as a function of the $\ell_2$-norm distance $d = \|\boldsymbol{\mu}-\boldsymbol{\mu}^*\|_2$ for $\boldsymbol{\mu}, \boldsymbol{\mu}^*\in\mathcal{P}$:
\begin{align}
    r_\text{Matérn}(d) = \varrho \, \frac{2^{1-\nu}}{\gamma(\nu)}\left(\frac{\sqrt{2\nu}\,d}{\lambda}\right)^{\nu}\,\mathcal{B}_\nu \left(\frac{\sqrt{2\nu}\,d}{\lambda}\right),
    \label{eq:matern_def}
\end{align}
where $\gamma(\cdot)$ is the Gamma function and $\mathcal{B}_\nu(\cdot)$ is a modified Bessel function, and where $\varrho$ is the signal variance parameter, $\lambda>0$ is the length-scale (or stability hyperparameter) and $\nu>0$ is the smoothness hyperparameter. 

\paragraph{Smoothness hyperparameter}

The stochastic Gaussian process resulting from a Matérn kernel is $\lceil\nu\rceil-1$ times differentiable in the mean-square sense. The smoothness parameter $\nu$ will take the form $\nu=p+1/2,~p\in\mathbb{N}$, since it is a common choice in machine learning framework (e.g.~\citeauthor{elbeltagi2021prediction}, \citeyear{elbeltagi2021prediction}; \citeauthor{mukesh2021prediction}, \citeyear{mukesh2021prediction}). Strong smoothness is irrelevant when dealing with experimental data. It might be hard to distinguish high values of smoothness $\nu\ge 7/2$ (existence of high-order derivatives) from noisy data~\citep{williams2006gaussian}. In this work, since the training dataset is assumed to be noisy, we consider $\nu = 5/2$.

\paragraph{Length-scale hyperparameters}

The length-scale parameter $\lambda$ represents the level of variability in the reduced coefficients as a function of distance in the input space $\mathcal{P}$.  In this work, we consider different length-scales for each input parameter since the input parameters are of different nature (source location versus atmospheric inflow conditions). Anisotropy may be embedded using a distinct correlation length-scale per dimension:
\begin{align}
    d(\boldsymbol{\mu}^{(m)},\boldsymbol{\mu}^{(n)}) = \sqrt{(\boldsymbol{\mu}^{(m)}-\boldsymbol{\mu}^{(n)})^T\,\boldsymbol{\Lambda}\,(\boldsymbol{\mu}^{(m)}-\boldsymbol{\mu}^{(n)})},
\end{align}
where $\boldsymbol{\mu}^{(m)}$ and $\boldsymbol{\mu}^{(n)}$ are realizations of the input vector $\boldsymbol{\mu}$, and where $\boldsymbol{\Lambda}\in\mathbb{R}^{4\times 4}$ corresponds to the length-scale matrix. In this work, this matrix will be of the form $\boldsymbol{\Lambda}=\mathrm{diag}(1/(\lambda_{u_{z_c}}^2), 1/(\lambda_{z_0}^2), 1/(\lambda_{x_{src}}^2), 1/(\lambda_{z_{src}}^2))$, which assumes independent length-scales and discriminates Gaussian process instability according to each input dimension. This form of the Gaussian process length-scale matrix is referred to as automatic relevance determination (ARD) in the literature.

\subsubsection{Hyperparameter optimization}
\label{sec:gp_optim_procedures}

Hyperparameter settings have a substantial impact on the GPR model prediction performance. An optimization process is usually used to determine an optimal value for the hyperparameters rather than simply specifying them. In this study, we optimize the hyperparameters for each of the $L$ GPR models to adapt to the characteristic length-scale of each POD mode (Sect.~\ref{sec:gp_definition}). For this purpose, we compare different optimization approaches, both in terms of accuracy and efficiency.  

For a noisy GPR framework with the anisotropic Matérn kernel, the set of hyperparameters $\boldsymbol{\theta}_l$ includes the correlation length-scales, the noise variance as well as the Gaussian process variance, meaning that $\boldsymbol{\theta}_l = \{s_l^2,\varrho,\lambda_{u_{z_c}},\lambda_{z_0},\lambda_{x_{src}},\lambda_{z_{src}}\} \in \mathbb{R}^{6}$. Empirical Bayesian maximization is used to determine the optimal set of the hyperparameters maximizing their posterior:
\begin{equation}
\begin{aligned}
    \boldsymbol{\theta}_{l,\mathrm{opt}} & = \mathrm{argmax}_{\boldsymbol{\theta}_l} \, \log \, p(\boldsymbol{\theta}_l\,\lvert\, \mathcal{U},\mathcal{K}_l) \\
    \Rightarrow \boldsymbol{\theta}_{l,\mathrm{opt}} & = \mathrm{argmax}_{\boldsymbol{\theta}_l} \, \log \, p(\mathcal{K}_l \, \lvert \, \mathcal{U},\boldsymbol{\theta}_l) + \log \, p(\boldsymbol{\theta}_l), 
    \label{eq:posterior_hyperparams_good}
\end{aligned}
\end{equation}
The first term called the marginal log-likelihood is assumed to be Gaussian; and the second term involves the prior distribution over the hyperparameters. In the literature, gradient descent is widely used to find the local optimum of Eq.~\ref{eq:posterior_hyperparams_good}~\citep{williams2006gaussian}.

\paragraph{Maximum log-likelihood estimation}

Without making any further assumption about the noise or the length-scales, one may proceed with a naive optimization of the log-likelihood, assuming uniform prior distributions over the hyperparameters (i.e. the term $p(\boldsymbol{\theta}_l)$ in Eq.~\ref{eq:posterior_hyperparams_good}). Unfortunately, gradient descent algorithms perform poorly in this case due to multiple local optima. To overcome this issue, one way is to perform multiple gradient descent iterations starting from different hyperparameter initial conditions. The final solution is then chosen as the one achieving the highest maximal log-likelihood (MLL) score. These multiple gradient descent iterations increase the computational cost of the GPR (recall that in this work we have $L$ GPR models to train and therefore we have to repeat this optimization procedure $L$ times).

\paragraph{Maximum a posteriori estimation} 

To ensure convergence to a solution consistent with reduced-basis properties, the optimization procedure can be informed by providing prior distributions (the term $p(\boldsymbol{\theta}_l)$ in Eq.~\ref{eq:posterior_hyperparams_good}) and an appropriate starting point for the hyperparameters. We propose in this work to infer these information from the POD coefficients (Sect.~\ref{sec:hyperparam_prior}). We refer to this approach as the maximum a posteriori (MAP) estimation. The MAP approach will be compared to the standard MLL approach in the following.

\subsection{Reduced-order model performance evaluation}

\subsubsection{Training, calibration and test}
\label{sec:dataset}

In this study, LES data (made of 750 snapshots -- Sect.~\ref{sec:sampling_strategy}) are split into three subsets following Halton's sequence ordering: \emph{i)}~a training dataset to learn the POD modes and the POD reduced coefficients using GPR (63\%, i.e. $N_{\text{train}} = 472$); \emph{ii)}~a calibration dataset to estimate GPR hyperparameter prior distributions (7\%, i.e. $N_{\text{calib}} = 53$); and \emph{iii)}~a test dataset to evaluate the capacity to predict LES quantities of interest for new samples of the uncertain input parameters (30\%, i.e. $N_{\text{test}} = 225$). Having a calibration dataset that is independent to the proper training dataset is essential to provide unbiased estimation of the Gaussian process prior distributions.

\subsubsection{Performance metrics}

\paragraph{Individual Gaussian process regression model}

In this work, we quantify the individual GPR model performance using a $Q^2$ (explained variance) criterion for each POD reduced coefficient. With this criterion, the $l$th GPR prediction error is weighted by the variance over the coefficients $\boldsymbol{k}_l$ of the $l$th POD mode:
\begin{align}
\label{eq:gpr_q2}
    Q^2_l &= 1-\frac{\displaystyle \|\boldsymbol{k}_l - \boldsymbol{m}_l^* \|^2_2}{\displaystyle \| \boldsymbol{k}_l - \hat{\mathbb{E}}[\boldsymbol{k}_l] \|^2_2} \, ,~\forall l=1,\,\hdots\,,L,
\end{align}
where $\boldsymbol{m}_l^*$ is the $l$th GPR mean prediction (Eq.~\ref{eq:posterior_gpr2}). In this study, $Q^2_l$ is estimated over the test dataset. Since we use whitening, reduced coefficients have zero-mean and unit-variance, meaning that this per-mode $Q^2$ metric directly reflects the mean squared error (MSE) on the predicted reduced coefficients (i.e. $Q^2 \approx 1 - \text{MSE}$).

\paragraph{Reduced-order model}
To quantify the performance of the reduced-order model in the physical space, we also evaluate the $Q^2$ criterion on each feature. The reconstruction/prediction error is weighted by the variance over the LES samples for each grid element $i$:
\begin{align}
\label{eq:local_q2}
    Q^2_i &= 1-\frac{\displaystyle \|\mathbf{K}_{\mathrm{les},i} - \mathbf{K}_{\mathrm{rb},i} \|^2_2}{\displaystyle \| \mathbf{K}_{\mathrm{les},i} -  \hat{\mathbb{E}}[\mathbf{K}_{\mathrm{les},i}] \|^2_2} \, ,~\forall i=1,\,\hdots\,,N_h.
\end{align}
In this study, the $Q^2_i$ criterion is estimated over the LES training dataset for verification as well as over the LES test dataset for evaluating the reduced-order model prediction capacity. To help with the analysis, we also derive a global score from the variance weighted local $Q^2$ criterion as:
\begin{equation}
\begin{aligned}
    \displaystyle Q^2 & = \sum_{i=1}^{N_h} \omega_i\,Q^2_i, 
    \quad \omega_i = \frac{\displaystyle\hat{\mathbb{V}}(\mathbf{K}_{\mathrm{les},i})}{\displaystyle\sum_{j =1}^{N_h}\hat{\mathbb{V}}(\mathbf{K}_{\mathrm{les},j})},
\end{aligned}
\label{eq:global_q2}
\end{equation}
where $\hat{\mathbb{V}}(\mathbf{K}_{\mathrm{les},i})$ corresponds to the variance unbiased estimation over the snapshots defined in Eq.~\ref{eq:var_mean_ensemble}. This weighted average matches the usual explained variance criterion of the POD, in agreement with the problem of $\ell_2$-matrix norm maximization:
\begin{align}
Q^2 = \displaystyle\sum_{l=1}^L\,\sigma_l \bigg/ \displaystyle\sum_{l=1}^{N_h}\,\sigma_l.
\label{eq:explainedvariance}
\end{align}

\subsubsection{Learning procedure}

Building our POD/GPR reduced-order model can be summarized as the following steps.

\paragraph{Training stage (learning)}
    \begin{enumerate}
        \item Extract the POD modes (Eq.~\ref{eq:pod_diagonalization}) and truncate the reduced basis to the first $L$ modes (Eq.~\ref{eq:explainedvariance}) from the training data: $\lbrace \boldsymbol{\psi}_l\rbrace_{l = 1,\hdots,L}$, 
        \item Calibrate the GPR hyperparameter prior distribution from POD: 
        $\lbrace \boldsymbol{\theta}_l\rbrace_{l = 1,\hdots,L}$,
        \item Estimate the GPR hyperparameter posterior distribution using either MLL or MAP (Eq.~\ref{eq:posterior_hyperparams_good}): $\lbrace \boldsymbol{\theta}_{l,\text{opt}}\rbrace_{l = 1,\hdots,L}$.
\end{enumerate}

\paragraph{Prediction stage (validation)}
    \begin{enumerate}
        \item Compute the POD reduced coefficients $\lbrace k_l(\boldsymbol{\mu}^*)\rbrace_{l = 1,\hdots,L}$ for the test sample of input parameters $\boldsymbol{\mu}^*$ (Eqs.~\ref{eq:posterior_gpr}--\ref{eq:posterior_gpr2}),
        \item Perform inverse POD to recover the predicted mean tracer concentration fields $\mathbf{K}_{\mathrm{rb}}^*$ from the POD coefficients for the test sample (Eq.~\ref{eq:pod_inverse}),
        \item Compare the emulated fields with the reference LES test samples $\mathbf{K}_{\mathrm{les}}^*$ (Eqs.~\ref{eq:local_q2}--\ref{eq:global_q2}).
\end{enumerate}
In this work, POD is implemented using the randomized truncated singular value decomposition~\citep{halko2009finding} from the scikit-learn library~\citep{scikit-learn}. As for GPR, the MLL approach is implemented using scikit-learn, while the MAP approach is implemented using GPyTorch~\citep{gardner2018gpytorch}. 

\section{Results}
\label{sec:res}

In the following, we analyze POD modes and quality of representation (Sect.~\ref{sec:res_pod}), give details about how to inform GPR models using POD (Sect.~\ref{sec:hyperparam_prior}), and analyze the POD/GPR reduced-order model accuracy on mean tracer concentration fields using the full dataset (Sect.~\ref{sec:res_gp}). We then carry a numerical analysis of the reduced-order model robustness with smaller training dataset (Sect.~\ref{sec:res_robust}). We end this section with an extension of the reduced-order model to the turbulent scalar flux (Sect.~\ref{sec:res_fluctu}).

\subsection{Analysis of proper orthogonal decomposition}
\label{sec:res_pod}

This section investigates the quality of representation of the ensemble variance as well as the mode spatial structures for time-averaged quantities of interest. We explain the choices that lead to the selection of the number of POD modes $L$. The number of POD modes directly determines the number of GPRs to perform (Eq.~\ref{eq:rom}).

\paragraph{Cumulative explained variance}

\begin{figure}[hbt]
\centering
\includegraphics[width=0.7\textwidth]{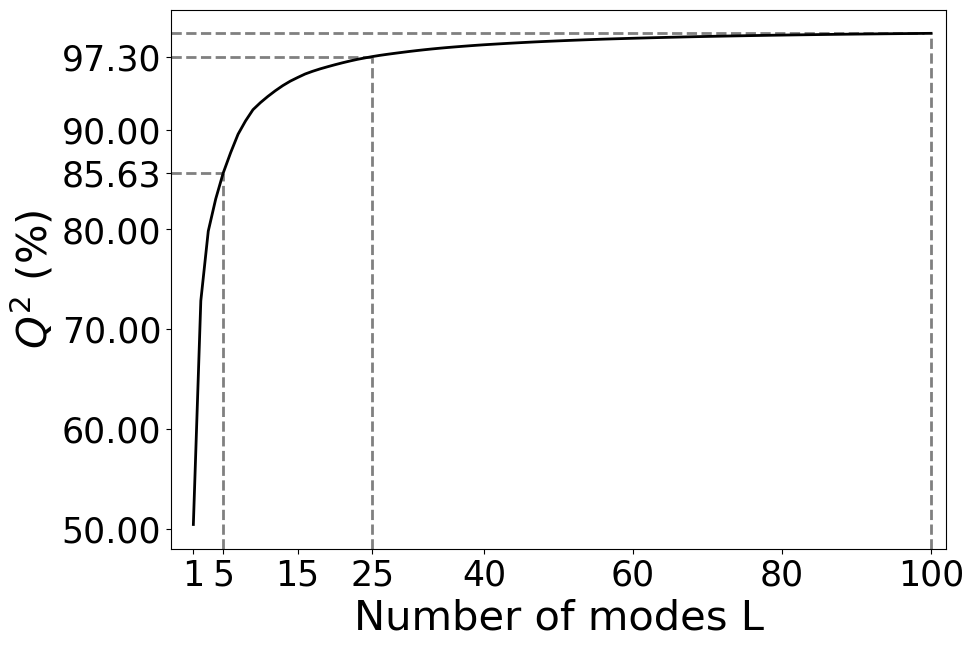}
\caption{Cumulative explained variance ($Q^2$ in \%, see Eq.~\ref{eq:explainedvariance}) for the POD reduced-basis size $L$ varying between 1 and 100 (solid line). The truncation thresholds for the Kaiser rule ($ L = 5$) and the elbow rule ($L = 25$) are represented in dashed lines.}
\label{fig:scree_graph}
\end{figure}

Several criteria are available in the literature for selecting the number of modes, the vast majority relying on the explained variance metrics evaluated on training data (e.g. the cumulative explained variance, the Kaiser and elbow rules -- \citeauthor{jolliffe2002springer},~\citeyear{jolliffe2002springer}). Figure~\ref{fig:scree_graph} shows the cumulative explained variance of the training dataset (Sect.~\ref{sec:dataset}) decomposed on the POD reduced basis as a function of the $L$ POD modes. In our case, the first mode alone contributes to about 50\% of the explained variance, the first fifteen modes explain more than 95\%. Alternatively, the Kaiser rule applied to 70\% of the mean eigenvalue suggests keeping only twenty-five modes; this corresponds to 97.3\% of the total ensemble variance. The elbow rule estimates the truncation threshold from the sign change in the eigenvalue second-order derivative: only the first fives modes shall be kept following this rule, which corresponds to 85.6\% of the total ensemble variance. 
This analysis shows that these different approaches lead to very different truncation choices. A finer analysis of the POD modes is necessary to determine an appropriate truncation level $L$.

\paragraph{Mode interpretation}

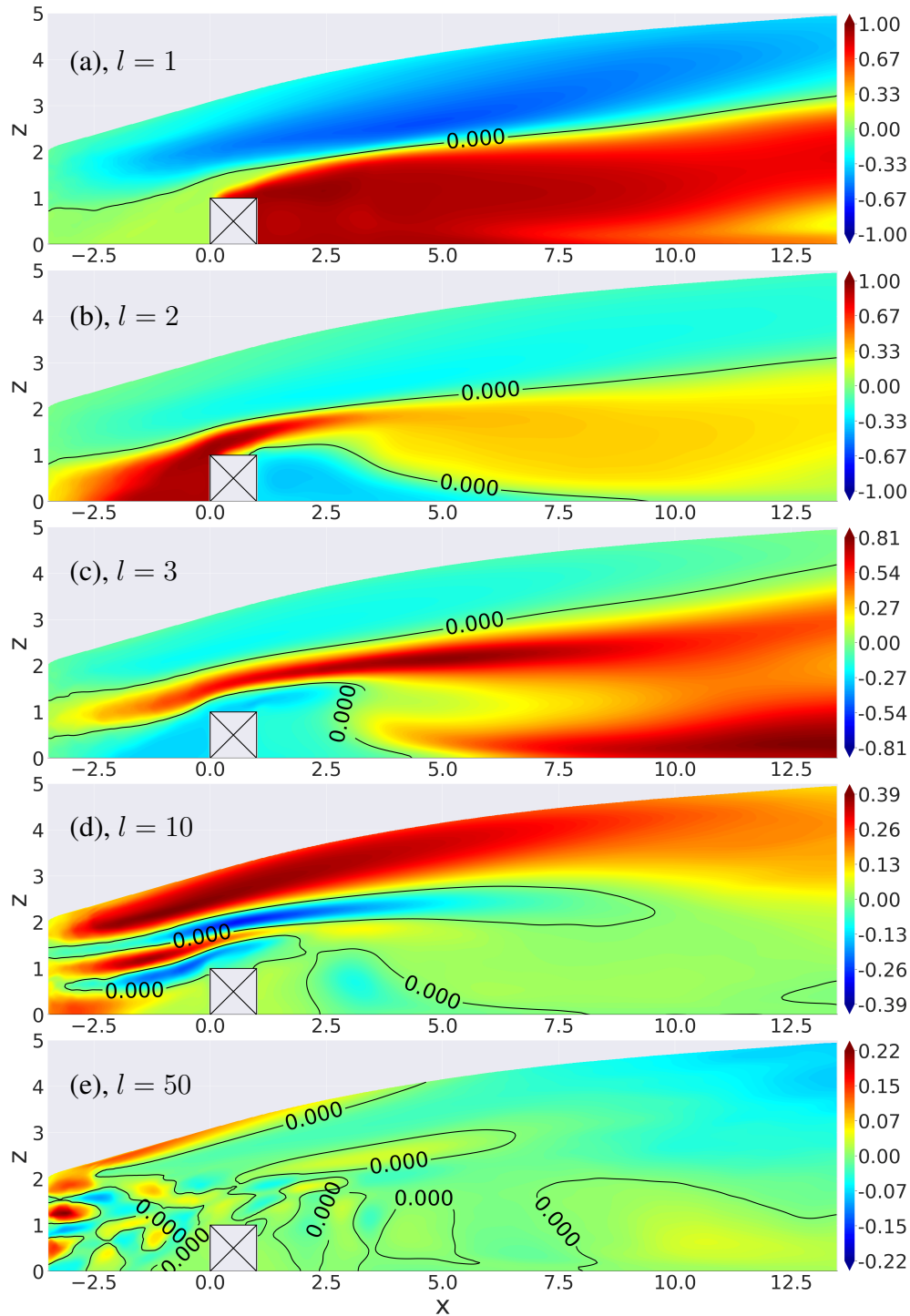
\begin{figure}[!htb]
\centering
\begin{overpic}[width=0.8\textwidth]{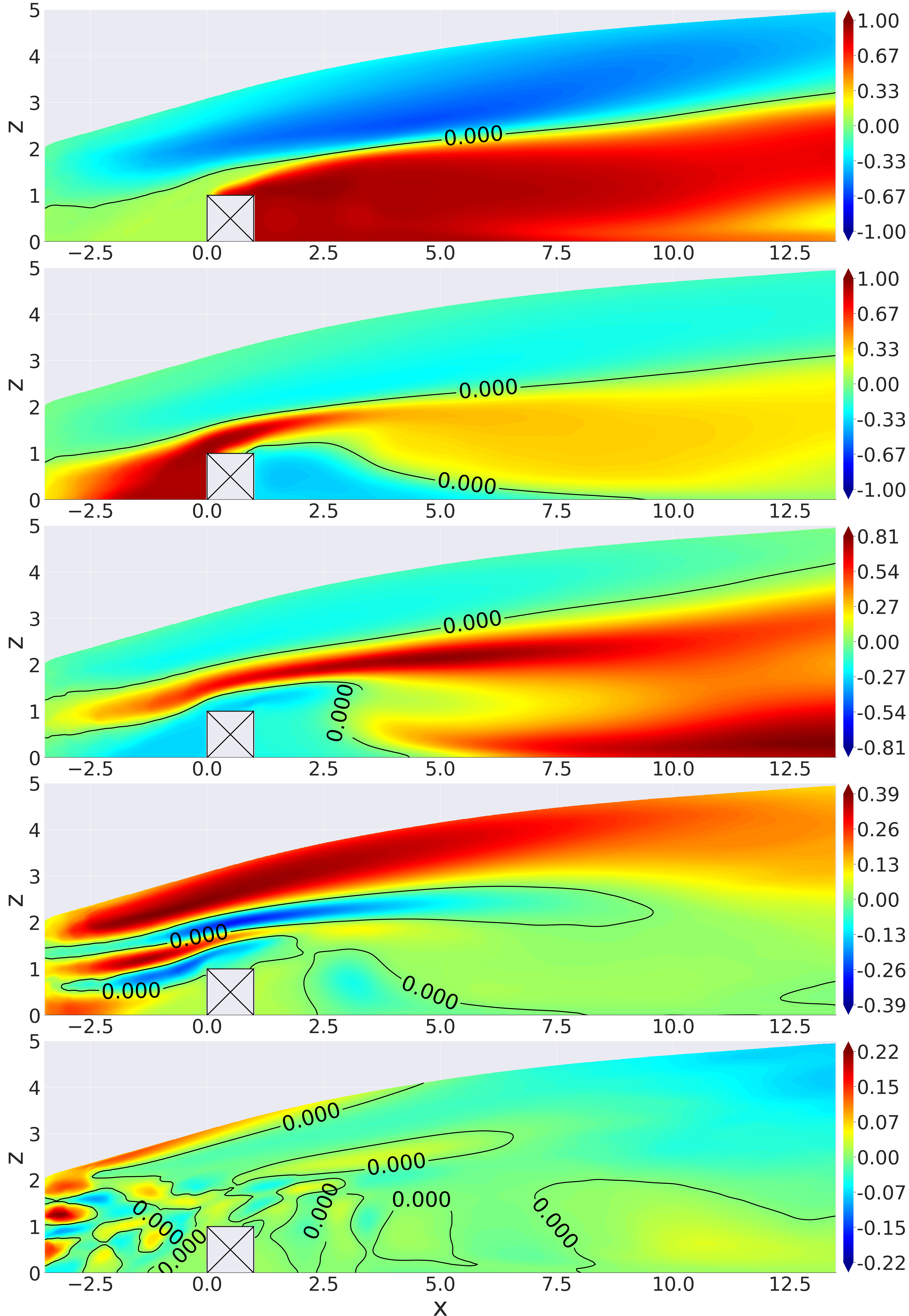}
 \put (5,95) {\large(a), $l = 1$}
 \put (5,75.5) {\large(b), $l = 2$}
 \put (5,56) {\large(c), $l = 3$}
 \put (5,36.5) {\large(d), $l = 10$}
 \put (5,16.8) {\large(e), $l = 50$}
\end{overpic}
\caption{Correlation maps (between -1 and 1) between the $l$th POD mode $\boldsymbol{\psi}_l$ and the LES snapshots (Eq.~\ref{eq:pod_correlation}) sorted by increasing order of POD mode indices from $l = 1$ to $l = 50$ (i.e. by decreasing order of eigenvalues $\sigma_l$).}
\label{fig:pod_modes}
\end{figure}

We now analyze the variance structures carried by the POD modes from the correlation maps in the physical space (Eq.~\ref{eq:pod_correlation}). Figure~\ref{fig:pod_modes} presents five out of the first hundred POD modes, which carry patterns that are representative of the whole set of POD modes. The first modes have spatially widespread structures with horizontally elongated shapes, looking like streaks aligned with the streamwise direction.

The correlation structures are associated with physical patterns of the mean LES fields. The first POD mode (Fig.~\ref{fig:pod_modes}a) consists of two anti-correlated horizontal layers. Large tracer concentrations in the lower part of the domain are unlikely to occur along with high concentrations in altitude because of the flow horizontal structure. Indeed, the plume dispersion is more or less affected by the obstacle depending on the emission source location. For the nominal snapshot (Fig.~\ref{fig:mean_nominal}a), we observe that the plume dispersion is primarily controlled by the vortex shedding resulting from the flow interaction with the obstacle and the development of a recirculation region downstream of the obstacle. The corner eddy enhances the accumulation of tracer concentration upstream of the obstacle, while downstream tracer concentration is dispersed by turbulence. Overall, in the nominal case, high tracer concentration values remain localized near the ground. Differently, when the emission source is positioned higher (see example of Fig.~\ref{fig:mean_nominal}c), the plume remains far from the ground and from the obstacle.

The second POD mode (Fig.~\ref{fig:pod_modes}b) highlights the variety of concentration fields within the ensemble, and shows interactions between plume and recirculation areas, depending if the tracer source is upstream or downstream of the obstacle. When the source is located upstream and sufficiently close to the ground, the obstacle constrains the plume dispersion in an accumulation area close to its left boundary. Similarly, when the emission source is located downstream close enough to the obstacle (see example of Fig.~\ref{fig:mean_nominal}b), the tracer is trapped in a second recirculation area where large tracer concentration values can be obtained. 

The third mode (Fig.~\ref{fig:pod_modes}c) highlights the area where the plume stops interacting with recirculation areas. This occurs when the emission source is located sufficiently far from the obstacle and from the ground. In that case, the plume is transported above the obstacle and disperses vertically further downstream because of vortex shedding.

From the tenth POD modes (Fig.~\ref{fig:pod_modes}d), we can see narrower correlation structures that are associated with the near-source advection-dominated dispersion of the tracer by the mean flow. The different streaks are directly related to the different source positions of the LES database since upstream emission source locations induce very refined, horizontally-elongated plume wakes.

Finally, the example of the fiftieth mode (Fig.~\ref{fig:pod_modes}e) highlights that the very high POD modes focus on the remaining ensemble variance heterogeneity. This mode features very localized bubble structures, which match tracer concentration peaks due to emission source locations present in the POD training database. The small structure values also raise the question of the potential noise induced in the POD modes due to the lack of training data volume.

\paragraph{Field prediction example}

\begin{figure}
\centering
\begin{overpic}[width=0.8\textwidth]{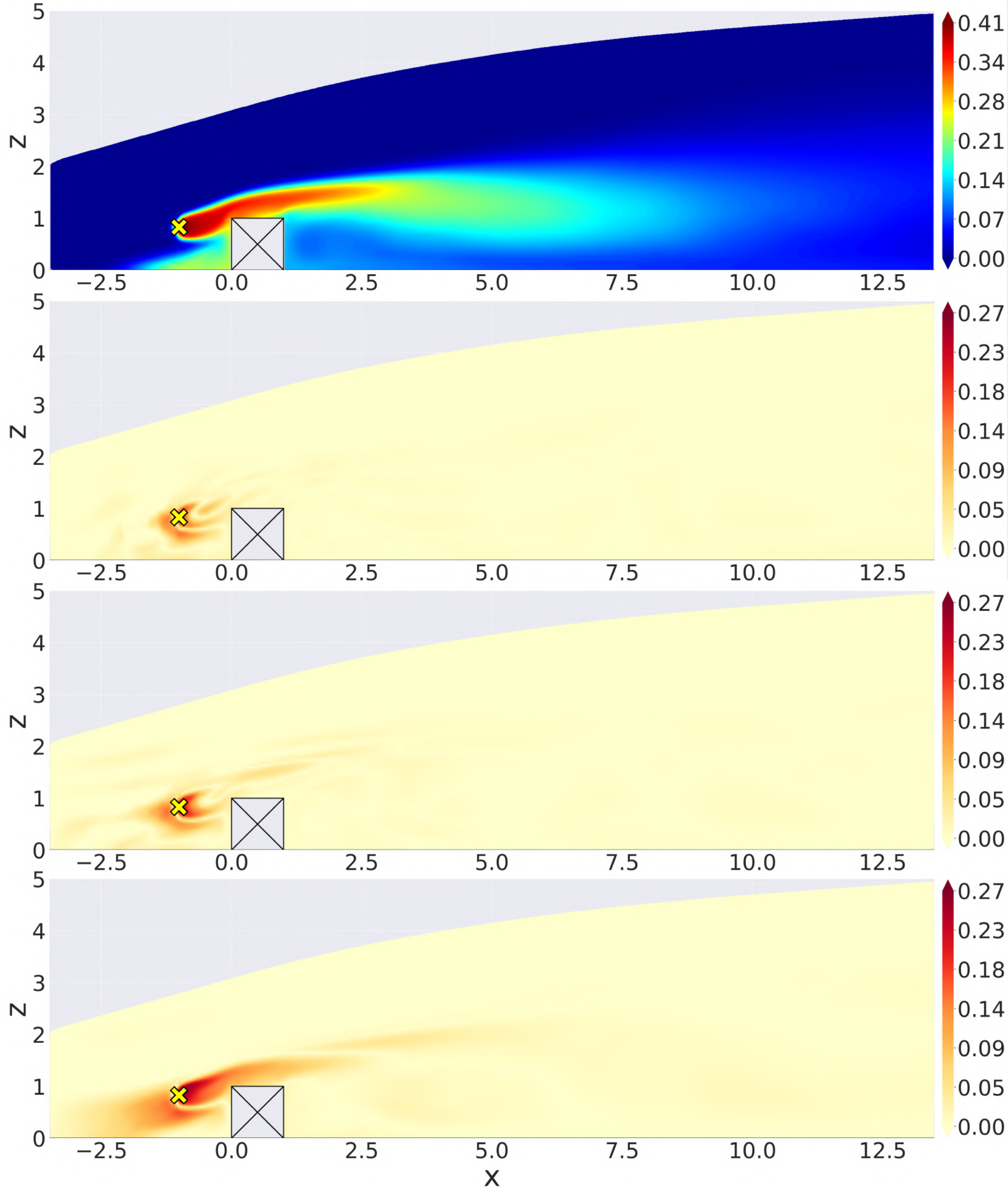}
 \put (5,94) {\large(a), LES}
 \put (5,69) {\large(b), $L = 100$}
 \put (5,44.5) {\large(c), $L = 26$}
 \put (5,20.5) {\large(d), $L = 5$}
\end{overpic}
\caption{Inverse POD for the nominal snapshot. (a)~LES reference solution (mean normalized tracer concentration field). POD reconstruction absolute error using (b)~$L = 100$ modes, (c)~$L = 26$ modes (Kaiser rule), and (d) $L = 5$ modes (elbow rule).}
\label{fig:nominal_pod_rebuilt}
\end{figure}
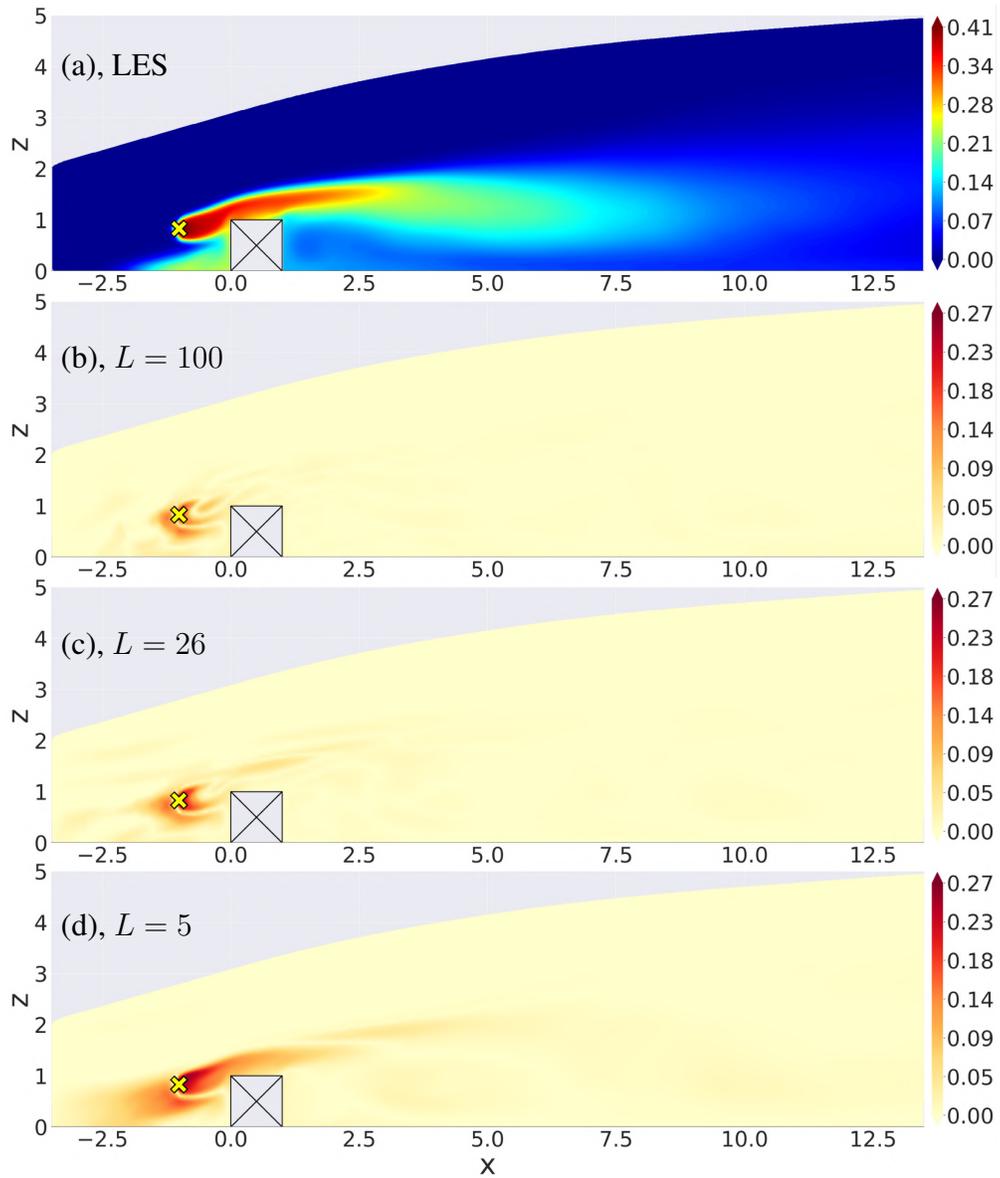

Figure~\ref{fig:nominal_pod_rebuilt} illustrates through the example of the nominal snapshot (Fig.~\ref{fig:mean_nominal}a) that is part of the LES test database, the capacity of reconstructing the mean tracer concentration fields from POD for different truncation levels $L$. Only accounting for the very first modes as suggested by the elbow rule ($L = 5$) provides a good representation of highly dispersed tracer areas downstream and of the tracer accumulation in the recirculation areas near the obstacle (Fig.~\ref{fig:nominal_pod_rebuilt}d). However, it does not handle well sharp patterns resulting from tracer advection-dominated dispersion near the emission source. Moreover, the peak emission of the source is underestimated and not well located. The Kaiser rule (Fig.~\ref{fig:nominal_pod_rebuilt}c) partially reconstructs the wake structures by including more modes ($L = 26$) but cannot correctly recover the plume structure close to the emission source. When including up to $L = 100$ modes in the reduced basis (Fig.~\ref{fig:nominal_pod_rebuilt}b), improvements in the field reconstruction mainly relate to the intensity and location of the source peak emission as well as to the localized structures around the emission source.

To conclude this section, high-order modes ($l \geq 50$ for instance) contain rich information on the near-source physics embedded in the LES snapshots. This large number of modes relates to the specificity of our problem, as changing the source location induces very sharp and localized plume structures near the emission source in the ensemble. For this reason, different snapshots might be very poorly correlated around the obstacle. Using POD standard approach, it is essential to keep a large number of modes to well represent local spatial concentration structures. We therefore choose at this stage, to keep $L = 100$ POD modes in the reduced basis that represent 99.6\% of the explained variance (Fig.~\ref{fig:scree_graph}). Using this reduced basis, we investigate how to efficiently tune a GPR to map POD coefficients of mean tracer concentration fields from any set of uncertain parameters.

\subsection{Gaussian process hyperparameter prior distribution}
\label{sec:hyperparam_prior}

GPR requires optimizing hyperparameters $\boldsymbol{\theta}_l = \{s_l^2,\varrho,\lambda_{u_{z_c}},\lambda_{z_0},\lambda_{x_{src}},\lambda_{z_{src}}\}$ for the $l$th reduced-order model, which is carried out by gradient descent on the marginal log-likelihood (MLL) or on the posterior distribution (MAP). In the case of MAP, prior distributions are no longer assumed to be uniform (Sect.~\ref{sec:gp_optim_procedures}). This section presents how to calibrate Gamma and Gaussian distributions for the hyperparameters from POD modes and reduced coefficient features that are required as input to the MAP optimization procedure.

\paragraph{Noise prior} 

POD often assumes that the variability of the low-order reduced coefficients is related to systematic behavior among the LES dataset, whereas the variability carried by the high-order reduced coefficients measure noise~\citep{jolliffe2002springer}. This is equivalent to treating the reduced coefficients on the first modes as unbiased data, which is a very restrictive assumption (Sect.~\ref{sec:meth_gpr}). In this study, GPR accounts for the noise on the $l$th reduced coefficient through the hyperparameter $s_l^2$ (Eq.~\ref{eq:eq_noise}). In this work, we assume that a prior estimate of $s_l^2$ may be obtained from the noise introduced by the time-averaging process performed on the LES (Sect.~\ref{sec:sampling_strategy}). Note that this is not the only source of noise introduced during the generation of the LES database, but this provides a lower bound noise estimation that is useful to demonstrate the added value of our methodology.

\begin{figure}[!htb]
\centering
\includegraphics[width=0.5\textwidth]{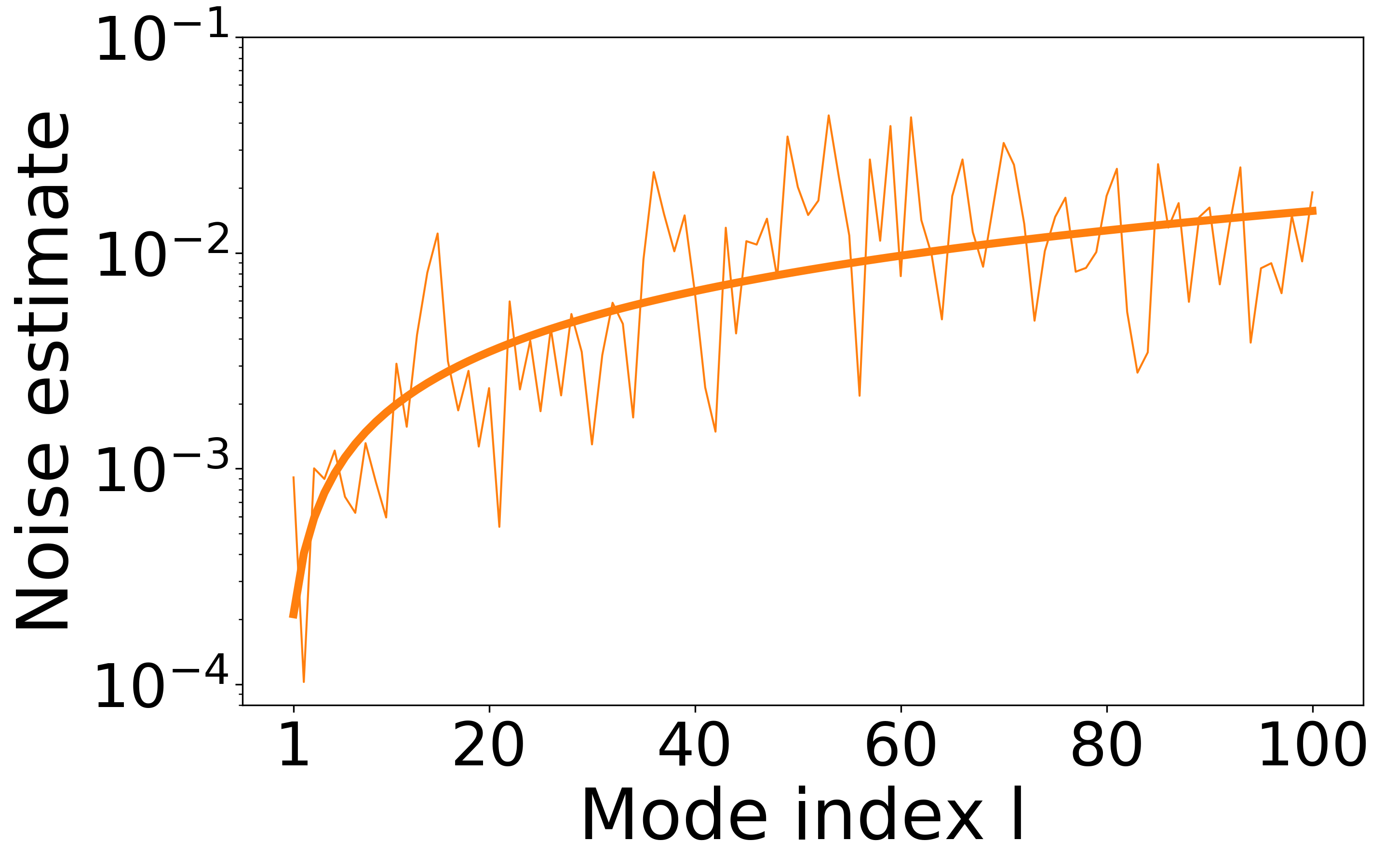}
\caption{Estimation of the noise hyperparameter $s_l^2$ for each reduced-order model $l$. The thin line corresponds to the noise estimation $\hat{s_l}^2$ from the POD modes (Eq.~\ref{eq:noise_estimator}). The thick line corresponds to the average trend found by regression and defined by the following equation $\hat{s}_l^2 = 2.16\times 10^{-4} \, l^{0.93}$.}
\label{fig:estim_noise}
\end{figure}

The main parameter involved in the time-averaging process is the length of the time-averaging window used to acquire LES statistics. To evaluate the noise for the $l$th reduced-order model, we compare the reduced coefficient of converged simulations ($k_l$) with the non-converged reduced coefficient obtained by averaging over only 50\% of the full simulation time window ($k_{l,50\%}$) as:
\begin{equation}
\label{eq:noise_estimator}
    \hat{s_l}^2 \approx \frac{1}{2N}\displaystyle\sum_{n=1}^N \left(k_l(\boldsymbol{\mu}^{(n)}) - k_{l,50\%}(\boldsymbol{\mu}^{(n)})\right)^2.
\end{equation}
Figure~\ref{fig:estim_noise} shows the $\hat{s_l}^2$-estimates obtained on the calibration samples (Sect.~\ref{sec:dataset}). The estimated noise increases from $10^{-4}$ to $10^{-2}$. It is small compared to the variability of the reduced coefficients (normalized to unity). Estimates of $s_l^2$ vary over several orders of magnitude and, since $s_l^2$ is positive, we set a Gamma prior distribution for $s_l^2$, whose mode matches the regressed estimate for each of the $L$ reduced-order models.

In addition, we assume POD has extracted most of the noise from the data so that noise magnitude tends to be small relatively to the total variability of the reduced coefficients. Since reduced coefficients are normalized to unit-variance, we have $ 0 \leq s_l^2 \leq 1$. The mean of the Gamma distribution is therefore set to 0.5 (middle value of the interval). For the hyperparameter optimization step, the starting point of the gradient descent for $s_l^2$ is taken as the Gamma distribution mode value that is obtained from a fit on the estimated noise (see thick line in Fig.~\ref{fig:estim_noise}).

\paragraph{Mean and scaling priors}

Now we establish prior information on the mean and variance of the Gaussian processes (Eq.~\ref{eq:gp_process}). In our GPR formalism, the mean is assumed to be constant, and the variance is decomposed into a systematic component and a noise component. From the expression of the noisy GPR models, we obtain for each POD reduced coefficient $l$: 
\begin{equation}
\mathbb{E}[k_l] = m_l \, , \quad \mathrm{Var}(k_l) = \varrho + s_l^2,
\end{equation}
where the signal variance hyperparameter $\varrho$ (Eq.~\ref{eq:matern_def}) characterizes systematic variability, while $s_l^2$ characterizes noise variability. $\varrho$ is linked to the successive transformations associated with POD centered and normalized reduced coefficients on the training dataset.

\begin{figure}[!htb]
\centering
\includegraphics[width=0.9\textwidth]{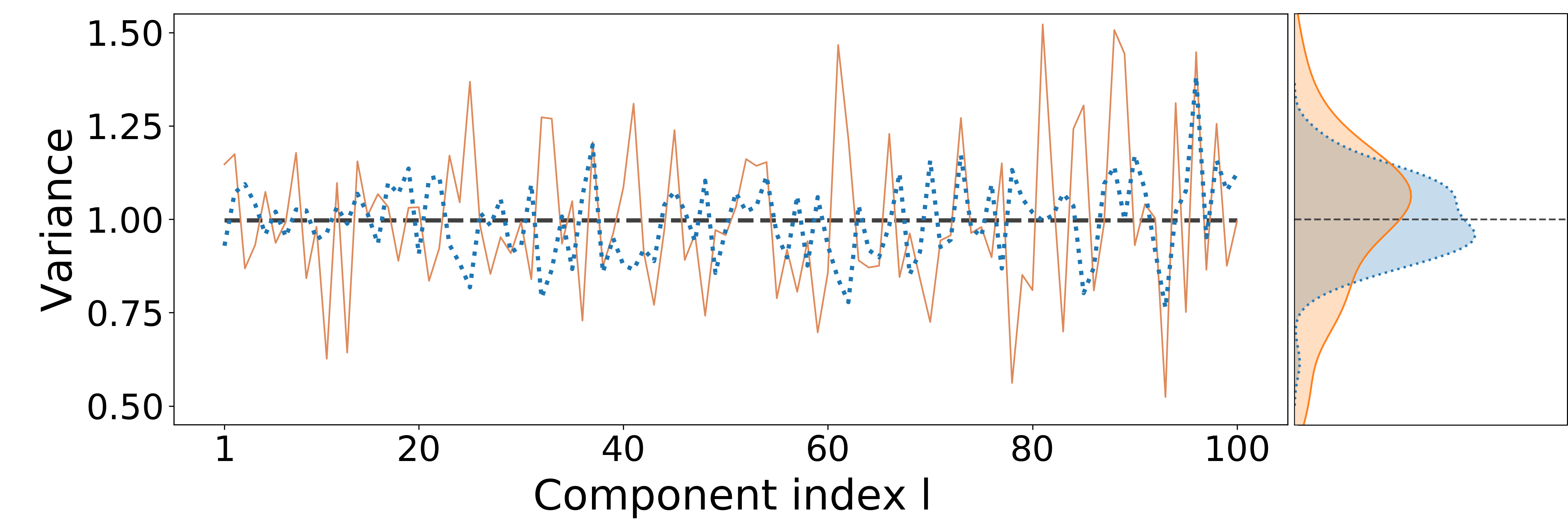}
\caption{Ensemble variance of reduced coefficients $\lbrace k_l\rbrace_{l = 1,\hdots,L}$ evaluated on the three subsets of the LES data: training data (dashed line), calibration data (solid line), and test data (dotted line). Related statistical distributions are plotted on the right.}
\label{fig:mean_std_distrib}
\end{figure}

Figure~\ref{fig:mean_std_distrib} shows the variance statistics of the reduced coefficients on the different LES datasets (Sect.~\ref{sec:dataset}). The variance evaluated on the training data is equal to one. This is no longer the case when the variance is evaluated on the calibration and test data. Still, the variance statistics oscillate on average around one with a Gaussian-like distribution. Similar behavior can be observed for the mean statistics that are equal to zero on the training data and on average around zero on the calibration and test data (not shown).

In this study, to be consistent with these results, the Gaussian process mean is set as deterministic with $m_l = 0$ for all $l = 1, \cdots, 100$. Concerning the variance, this is represented using a scaled Matérn kernel. When compared to the total variability of the Gaussian process, the proportion of variability attributed to noise (associated with $s_l^2$) appears to be small. For this reason, we assume that the total variability is essentially induced by systematic behavior of LES data and carried by the hyperparameter $\varrho$. We set a Gaussian prior distribution for $\varrho$ with mean and variance estimated from the calibration data ensemble statistics and set to 1 and 0.03, respectively. During the optimization step, the starting point of the gradient descent for $\varrho$ is set to the Gaussian distribution mean.

\paragraph{Correlation length-scale prior}

We are now interested in the prior distributions of the Matérn kernel correlation length-scales $\{\lambda_{x_{src}}, \lambda_{z_{src}}, \lambda_{U_{z_c}}, \lambda_{z_0}\}$ (Sect.~\ref{sec:gpr_kernel_spec}). The analysis of the POD modes (Fig.~\ref{fig:pod_modes}) reveals an increase in small-scale spatial heterogeneity for increasingly higher-order modes, which results from the ensemble variability driven by the emission source location. 
The first modes have spatially widespread structures with horizontally elongated shapes, looking like streaks aligned with the streamwise direction. When the POD mode index increases, the number of alternated streaks increases and these streaks get narrower along the vertical direction, while a distinct region emerges in the recirculation zone downstream of the obstacle.
Due to the alternated signs of the POD modes, it seems natural to anticipate the correlation lengths of the random processes used to model the POD coefficients as being strongly influenced by these patterns. In particular, we can see that the typical correlation length along the source height should be related to the number of streaks, and should therefore decrease as we consider higher POD modes. 
We therefore model a decrease in the length-scales $\lambda_{x_{src}}$ and $\lambda_{z_{src}}$ associated with $x_{src}$ and $z_{src}$ when moving to higher-order modes. 
For low-order POD modes, the length-scales tend to be large (i.e. the process is stable relatively to source position and height). For high-order POD modes, they tend to be smaller (i.e. the process is unstable relatively to source position and height). 
Because the correlation length-scales are positive, we adopt a prior Gamma distribution for $\lambda_{x_{src}}$ and $\lambda_{z_{src}}$. The mode of each Gamma distribution is determined by the simple decreasing rule: $1/l$ for $l = 1,\hdots, 100$. The variance of the Gamma distribution is set to 1, which corresponds to the interval length associated with the normalized input parameters ($[0,1]^4$, Sect.~\ref{sec:sampling_strategy}). Since there is no analogous interpretation of decreasing length-scales for $z_0$ and $U_{z_c}$, we retain prior Gamma distributions with constant mode and variance equal to 1 for $\lambda_{U_{z_c}}$ and $\lambda_{z_0}$. During the hyperparameters optimization step, the starting point of the gradient descent for $\{\lambda_{x_{src}}, \lambda_{z_{src}}, \lambda_{U_{z_c}}, \lambda_{z_0}\}$ is set to the Gamma distribution mode.

To conclude this section, it is possible to define prior distribution for the GPR hyperparameters based on POD information. This information will then be used as a starting point of the gradient descent in the MAP optimization step of GPR. 

\subsection{Gaussian process regression performance}
\label{sec:res_gp}

In this section, we compare the reduced-order model solutions obtained from MLL and MAP hyperparameter optimization procedures to show the added value of MAP. We also quantify the noise of the coefficients on each POD mode, and carry out a performance analysis using the $Q^2$ metrics.

\paragraph{Comparison of optimization solutions}

\begin{figure}[!htb]
\centering
\begin{overpic}[width=0.9\textwidth]{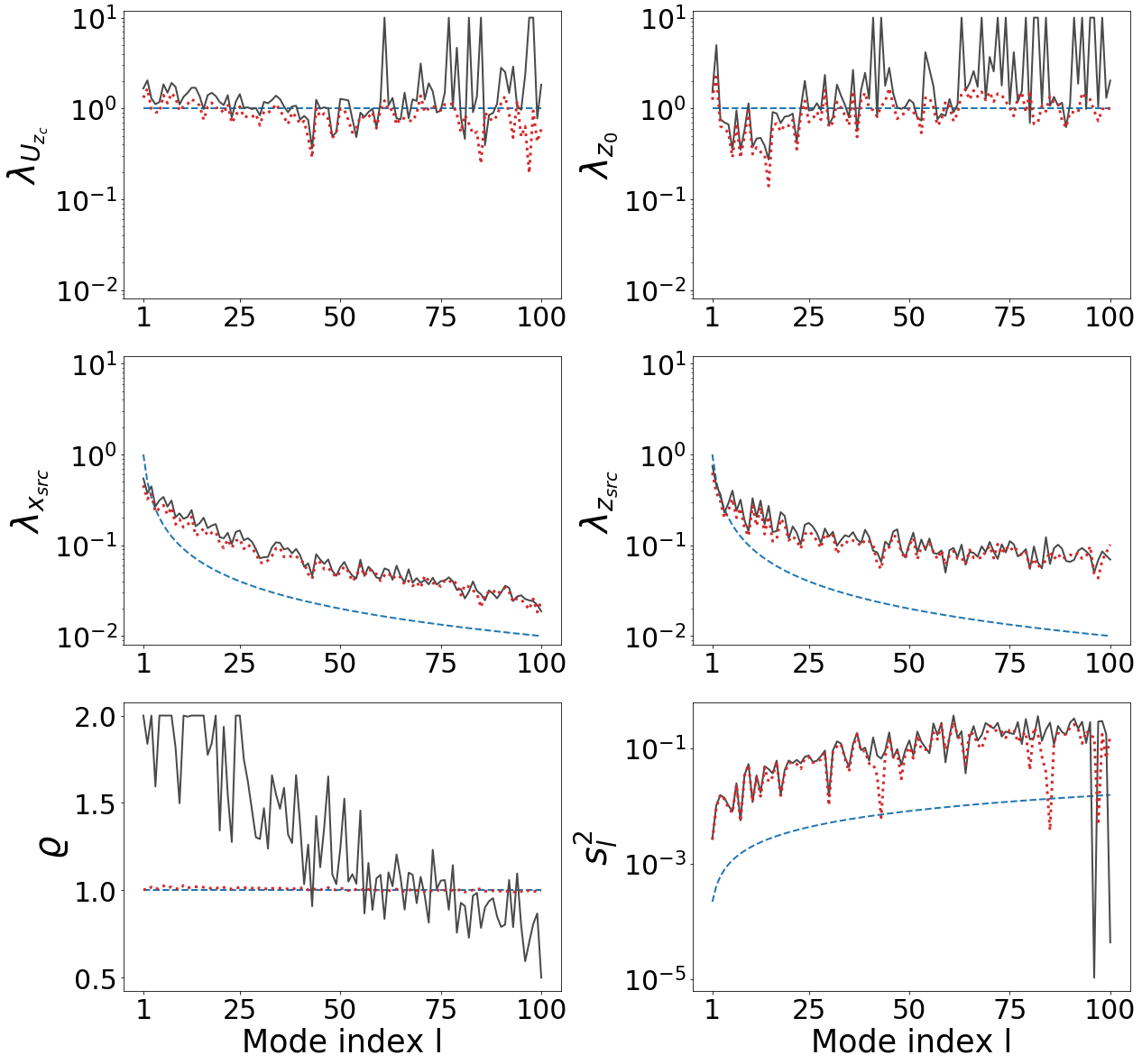}
 \put (13,70) {\large(a)}
 \put (62.5,70) {\large(b)}
 \put (13,39.5) {\large(c)}
 \put (62.5,39.5) {\large(d)}
 \put (13,9) {\large(e)}
 \put (62.5,9) {\large(f)}
\end{overpic}
\caption{Gaussian process correlation length-scales for (a)~$u_{z_c}$, (b)~$z_0$, (c)~$x_{src}$, and (d)~$z_{src}$. Correlation length-scales can be optimized using MLL (solid lines) or MAP (dotted lines) starting from prior modes (dashed lines). Noise in the MLL and MAP estimates (e) is used to determine $s_l^2$ (f).}
\label{fig:gp_lengthscales}
\end{figure}
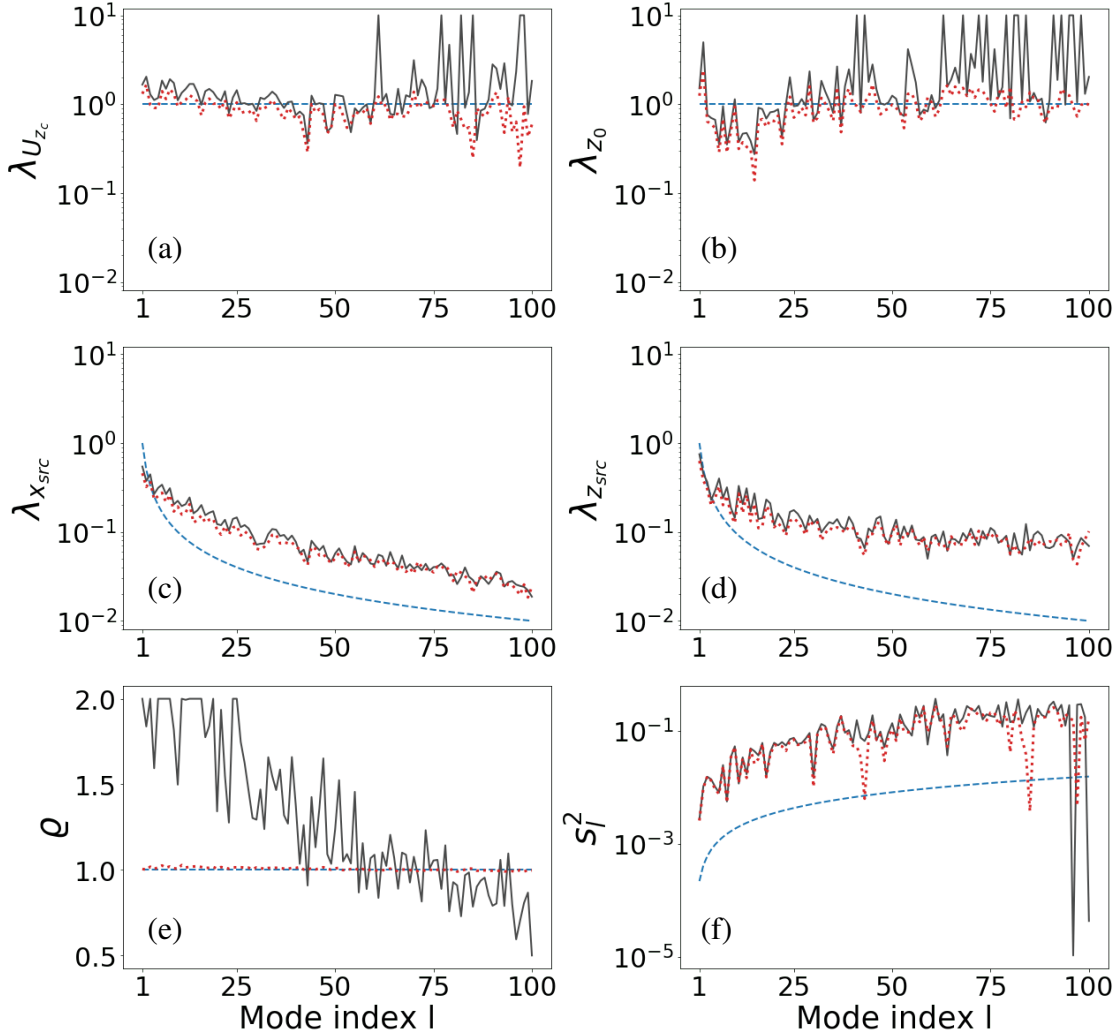

Figure~\ref{fig:gp_lengthscales} presents the modes of the hyperparameter prior distribution (dashed lines) as well as the optimized hyperparameter solutions obtained with MAP (dotted lines) to compare with MLL (solid lines). Recall that MAP takes as input the GPR hyperparameter prior information (Sect.~\ref{sec:hyperparam_prior}). Figure~\ref{fig:gp_lengthscales}abcd shows that MAP and MLL procedures converge to similar hyperparameter solutions. Estimated correlation length-scales are close to each other. However, MAP estimates are systematically underpredicted compared to MLL solutions on the first fifty modes. 

Noise estimates in Fig.~\ref{fig:gp_lengthscales}ef are consistent for almost all POD modes. Only the estimates for the variance hyperparameter $\varrho$ strongly differ between MLL and MAP procedures. To explain this difference, we note that the prior distribution associated with $\varrho$ features a low variance. Therefore, the MAP procedure converges to values close to the prior distribution mode. The MLL procedure is not constrained by the prior distribution, and the $\varrho$-estimates diverge on the first modes towards the upper bound value (set at~2). Additional tests (not shown) demonstrated that the MLL procedure applied with $\varrho=1$ leads to correlation length-scales identical to MAP estimates. This indicates that the $\varrho$-estimates obtained with MLL and MAP impact the associated length-scales. For MLL, the excess of variance on the low-order modes (associated with a larger value of $\varrho$) is balanced by a greater stability with respect to the input parameters (associated with larger correlation length-scales). From the fiftieth mode onwards, the MLL $\varrho$-estimates decrease around one, and the MLL correlation length-scales are no longer systematically greater than the MAP correlation length-scales. The decrease in $\varrho$ is explained by the fact that the variance on the reduced coefficients remains close to 1 for each POD mode (Sect.~\ref{sec:build_pod_dataset})  and at the same time $s_l^2$ increases (Fig.~\ref{fig:gp_lengthscales}).

In the end, the major difference between the MLL and MAP results concerns $\varrho$ on the first POD modes and consequent length-scales. Nevertheless, since noise is negligible on the first modes (because $\varrho \gg s_l^2$), the associated Gaussian process mean predictions are almost equal (Eq.~\ref{eq:posterior_gpr}). Hence, MLL and MAP procedures converge towards similar optima for the GPR hyperparameters, meaning that they correspond to equivalent predictive models but with an economy concerning numerical costs for MAP compared to MLL: the coarse prior distributions are sufficient to ensure MAP convergence in a single gradient descent when fifteen iterations are required for MLL (this factor of fifteen is all the more important as this optimization procedure is repeated for each of the $L = 100$ GPR models).

\paragraph{Gaussian process noise analysis}

Figure~\ref{fig:gp_lengthscales}f also shows that the estimated noise ($s_l^2$) is two orders of magnitude above the prior solution. This suggests that the temporal convergence error may not be the primary source of noise in the data. On the first POD modes, the noise is small compared to the signal variance. This is consistent with the fact that the first POD modes are almost noise free since they carry large variance structures. But the noise increases continuously on high-order modes such that it can be substantial compared to the signal variance (the maximum noise estimate is close to 0.3 for very high modes, which is of the same order of magnitude as the signal variance that is around 1). 

This gradual increase of noise indicates that there are not two distinct behaviors among the modes, namely those that carry systematic information on the one hand and those that carry noise on the other hand. The ratio between $s_l^2$ and $\varrho$ could be a way to choose the number of POD modes $L$ to keep in the reduced basis. 

\paragraph{Correlation length-scale analysis}

We now identify the most relevant input parameters in the GPR models. The order of magnitude for $\lambda_{U_{z_c}}$ and $\lambda_{z_0}$ is equivalent to $\lambda_{x_{src}}$ and $\lambda_{z_{src}}$ on the first POD modes (Fig.~\ref{fig:gp_lengthscales}abcd). This means that the variance on the reduced coefficients is equally due to the variations on all input parameters. On high POD modes, the values of $\lambda_{U_{z_c}}$ and $\lambda_{z_0}$ are larger than $\lambda_{x_{src}}$ and $\lambda_{z_{src}}$. This implies that the reduced coefficients are relatively insensitive to variations in $U_{z_c}$ and $z_0$, and are mainly explained by $x_{src}$ and $z_{src}$. This is why the optimization process for $\lambda_{U_{z_c}}$ and $\lambda_{z_0}$ becomes unstable for high-order POD modes. 

The correlation length-scales on position parameters $x_{src}$ and $z_{src}$ decrease in a stable manner over the first hundred POD modes as in the prior solutions. This suggests that the value of the reduced coefficients becomes more sensitive to small variations in source position and height on the higher POD modes. This is consistent with the small structures observed in the high-order modes and associated with tracer concentration wakes (Sect.~\ref{sec:res_pod}). A large number of POD modes ($L = 100$) is necessary here to characterize the high sensitivity of the tracer concentration upstream of the obstacle to the source height and position.

\paragraph{Gaussian process regression accuracy}

\begin{figure}[!htb]
\centering
\includegraphics[width=0.8\textwidth]{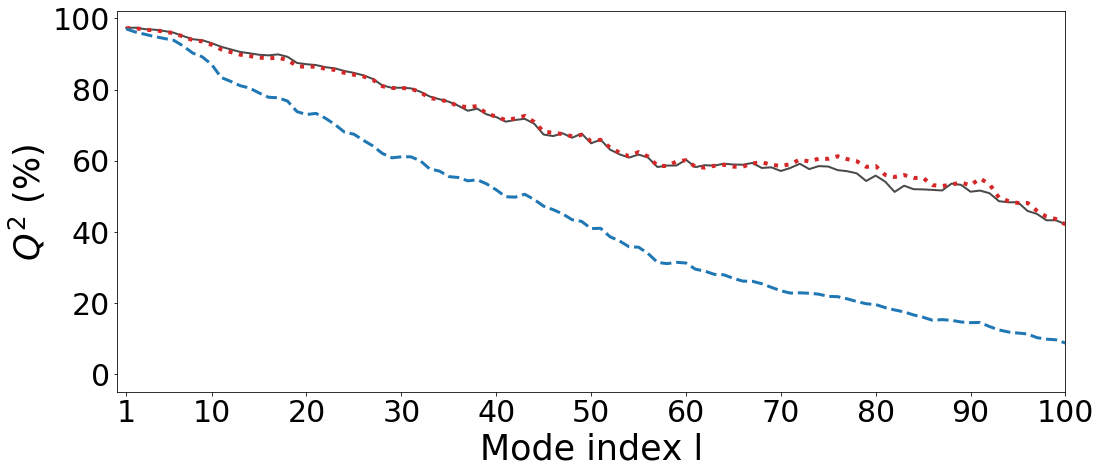}
\caption{Per mode-$Q^2$ (Eq.~\ref{eq:gpr_q2}) for GPR models with noise and length-scales that are either imposed using prior information (dashed line) or optimized by MLL (solid line) or MAP (dotted line).}
\label{fig:gp_q2}
\end{figure}

We now evaluate the accuracy of the GPR models over the test dataset (Sect.~\ref{sec:dataset}) for validation purpose. Figure~\ref{fig:gp_q2} shows the $Q^2$ scores obtained for each POD mode for the MLL and MAP procedures. For the very first modes, the $Q^2$ score is close to one, which means that GPR models perform almost perfectly on large variance structures with limited overfitting. However, they cannot maintain this level of performance when moving to high-order modes that feature more complex and localized structures. The decrease in performance appears linear, and the higher noisier modes are more difficult to predict (with a $Q^2$ coefficient evolving between 0.4 and 0.6 from the fiftieth mode onwards). 

Still, this adaptive procedure remains partly successful in predicting fine structures of the LES ensemble variance. Optimizing the hyperparameters mode per mode greatly improves the GPR accuracy compared to simply imposing prior trend on hyperparameters. This gain in $Q^2$ is more pronounced when moving to high-order modes. Results also confirm that the reduced-order models obtained from MLL and MAP procedures are equivalent as anticipated from the equivalent correlation length-scales in Fig.~\ref{fig:gp_lengthscales}. Globally, the $Q^2$ score of the reduced-order model evaluated on the test dataset is equal to 96.8\%. This model slightly overfits data since the $Q^2$ metric is equal to 99.3\% on the training dataset (recall that the reduced-order model cannot achieve a global $Q^2$ metric beyond 99.6\%, which is the upper limit imposed by the POD reconstruction error for $L = 100$ -- Fig.~\ref{fig:scree_graph}). Note that the $s_l^2$-hyperparameter slightly reduces accuracy on the training dataset but this is not of the same magnitude as the decrease in performance observed on the test dataset. This hyperparameter is therefore not responsible for the observed regression error. 

\paragraph{Field prediction examples}

\begin{figure}[!htb]
\centering
\begin{overpic}[width=0.9\textwidth]{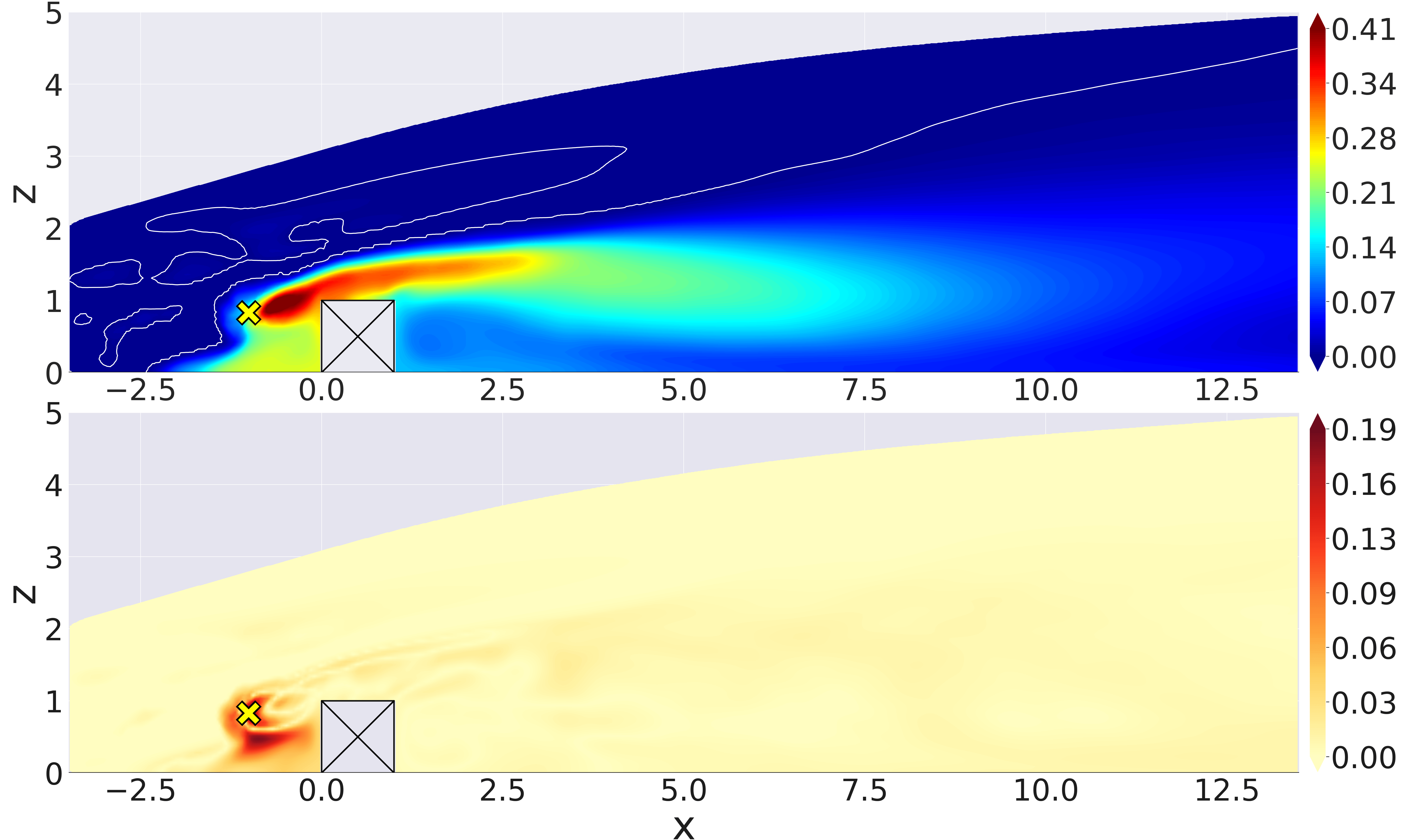}
 \put (7.5,53) {\large(a)}
 \put (7.5,24.5) {\large(b)}
\end{overpic}
\caption{Nominal snapshot mean normalized tracer concentration field obtained with (a)~reduced-order model prediction. (b)~Prediction absolute error calculated with respect to LES snapshot (Fig.~\ref{fig:mean_nominal}a). Contour line of the mean normalized tracer concentration equal to $5\times 10^{-4}$ is superimposed on reduced-order model predicted field to highlight the presence of low-magnitude noisy structures.}
\label{fig:nominal_concat}
\end{figure}

\begin{figure}
\centering
\begin{overpic}[width=0.9\textwidth]{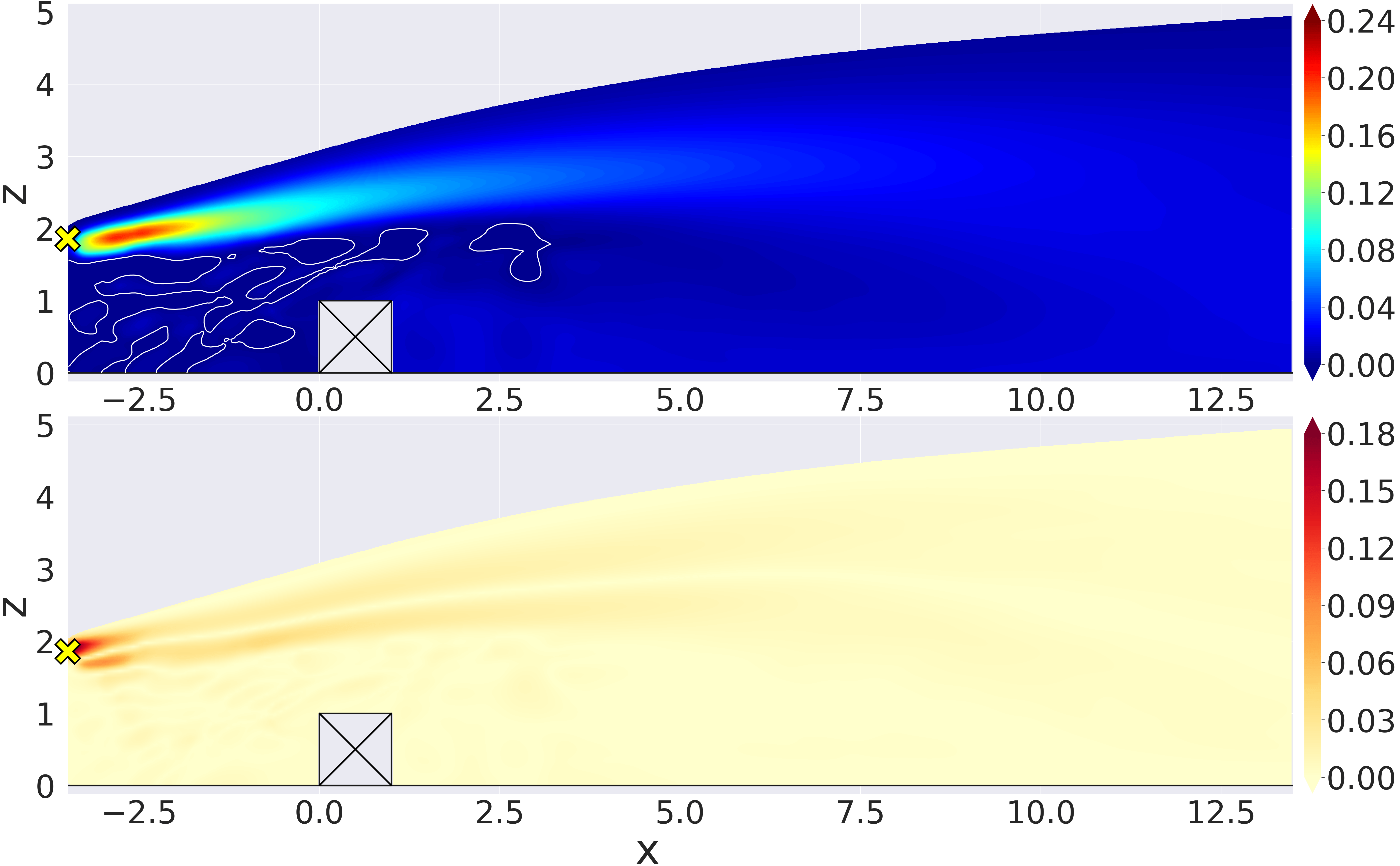}
 \put (7.5,54) {\large(a)}
 \put (7.5,25) {\large(b)}
\end{overpic}
\caption{Same caption as in Fig.~\ref{fig:nominal_concat} for a LES snapshot with a high emission source (Fig.~\ref{fig:mean_nominal}c).}
\label{fig:high_source}
\end{figure}

\begin{figure}
\centering
\begin{overpic}[width=0.9\textwidth]{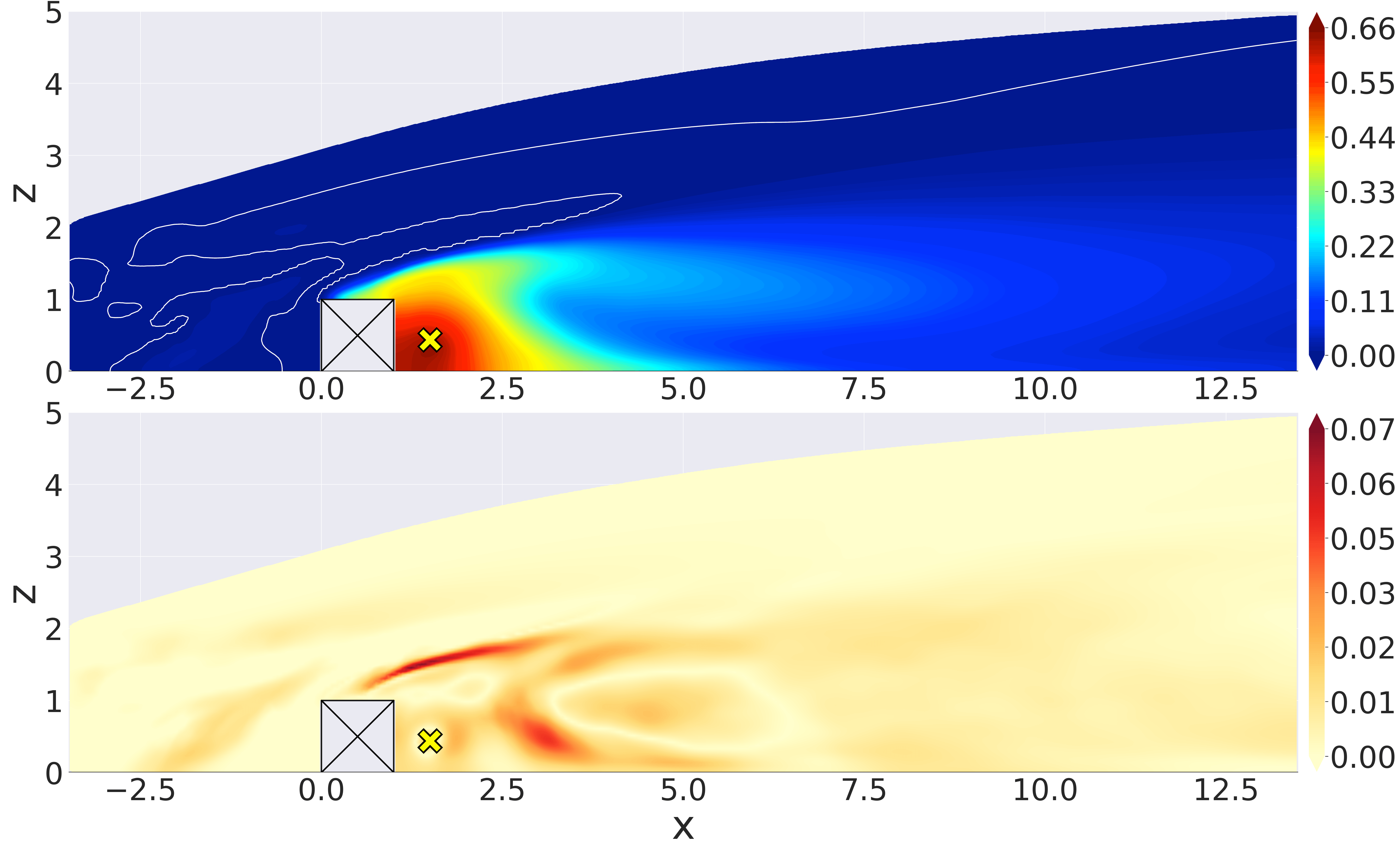}
 \put (7.5,53) {\large(a)}
 \put (7.5,24.5) {\large(b)}
\end{overpic}
\caption{Same caption as in Fig.~\ref{fig:nominal_concat} for a LES snapshot with an emission source downstream of the obstacle (Fig.~\ref{fig:mean_nominal}b).}
\label{fig:downstream_source}
\end{figure}

Figures~\ref{fig:nominal_concat}--\ref{fig:downstream_source} show the mean normalized tracer concentration fields predicted by the reduced-order model (resulting from MAP optimization) for three snapshots of the LES test dataset (Fig.~\ref{fig:mean_nominal}).

For the nominal snapshot (Fig.~\ref{fig:nominal_concat}), the largest prediction errors are made \emph{i)}~close to the emission source with largely underestimated tracer concentration (about 50\% of the LES reference concentration), and \emph{ii)}~in the accumulation area upstream of the obstacle with overestimated coarser-structured tracer concentration levels. 
Further away from the source, the tracer concentration in the wake of the obstacle is well predicted. Upstream of the obstacle, we can distinguish slight noise in no-tracer areas due to high-order POD modes that carry small noisy structures (Fig.~\ref{fig:pod_modes}).

We also present the prediction result for a case where the emission source is far from the obstacle and the ground (Fig.~\ref{fig:high_source}). This case emphasizes what has already been observed in the nominal case. The tracer concentration at the actual emission source and along the wake is hard to predict. The prediction absolute error can reach up to 75\% of the LES solution. The fine structures of the plume are hard to predict since information is mostly carried by multiple high-order POD modes in a region where there are only a few samples of the emission source in the LES dataset (this case is located at the boundary of the tracer source area -- Fig.~\ref{fig:scheme2D}).

To complement the analysis, we finally present the prediction result for a case where the emission source is located in the recirculation area downstream of the obstacle (Fig.~\ref{fig:downstream_source}). This case is much better predicted by the reduced-order model. The areas of high tracer concentrations form a wide-spread structure in areas carried by the first POD modes. Most of the information can be therefore conveniently recovered from the first reduced coefficients with accurate GPR models, including in the recirculation area downstream of the obstacle. The largest prediction error is on the order of 10\% of the LES solution.

These three snapshots highlight the capacity of the reduced-order model to predict the main tracer concentration structures, in particular when the emission source is located in recirculation zones near the obstacle. In this situation, the tracer concentration patterns are smooth as the dispersion is dominated by turbulent diffusion, making the prediction task easier. These results were obtained by considering a large number of modes in the reduced basis with a rich training dataset ($N_{\text{train}} = 472$). Such a large LES database is unaffordable in all realistic atmospheric dispersion cases involving larger computational domains and three-dimensional effects. We now evaluate how much the reduced-order model accuracy degrades when the size of the training dataset is reduced to have a better idea of the minimum required budget for building a multi-query uncertainty quantification framework.

\subsection{Robustness to training dataset}
\label{sec:res_robust}

We now reduce the training dataset to one hundred snapshots ($N_{\text{train}} = 100$) and fifty snapshots ($N_{\text{train}} = 50$) to move towards a more realistic multi-query LES framework. Note that 90\% of these training snapshots are used to determine the POD reduced basis, with the remaining 10\% used for calibration as before. This implies that the maximum number of modes in the reduced basis is directly equal to the POD training size (90 for $N_{\text{train}} = 100$; 45 for $N_{\text{train}} = 50$). The only difference is that here the whole dataset is used to optimize the GPR models since the dataset is of very limited size. Note also that the test dataset remains the same as before (as described in Sect.~\ref{sec:dataset}) to avoid introducing bias during the validation stage.

\begin{figure}[!htb]
\centering
\includegraphics[width=0.8\textwidth]{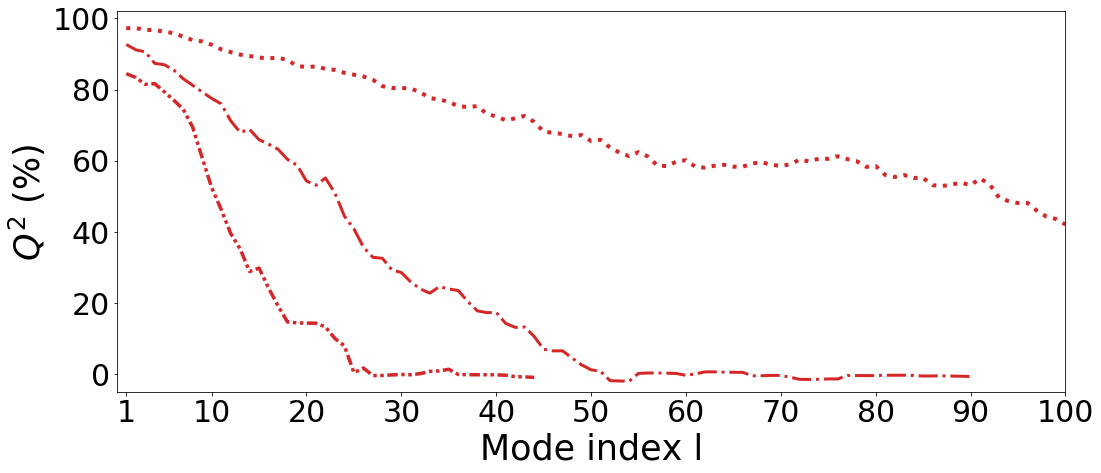}
\caption{Per mode-$Q^2$ (Eq.~\ref{eq:gpr_q2}) for GPR models trained with 472 snapshots (dotted line corresponding to the MAP result already presented in Fig.~\ref{fig:gp_q2}), 100 snapshots (dashdotted line) or 50 snapshots (densely dashdotted line).}
\label{fig:q2_per_mode_comp}
\end{figure}

Figure~\ref{fig:q2_per_mode_comp} compares the evolution of per-mode $Q^2$ for both full and reduced training datasets. The reduced-order model accuracy significantly decreases for the reduced database. There is a faster linear decrease towards $Q^2=0$ than with the full training dataset (the threshold $Q^2=0$ is approximately reached for the fiftieth mode for $N_{\text{train}} = 100$ and the the twenty-fifth mode for $N_{\text{train}} = 50$). This suggests that the number of POD modes to consider in the reduced-order models should be reduced due to the too limited size of the training dataset. 

\begin{figure}[!htb]
\centering
\includegraphics[width=0.9\textwidth]{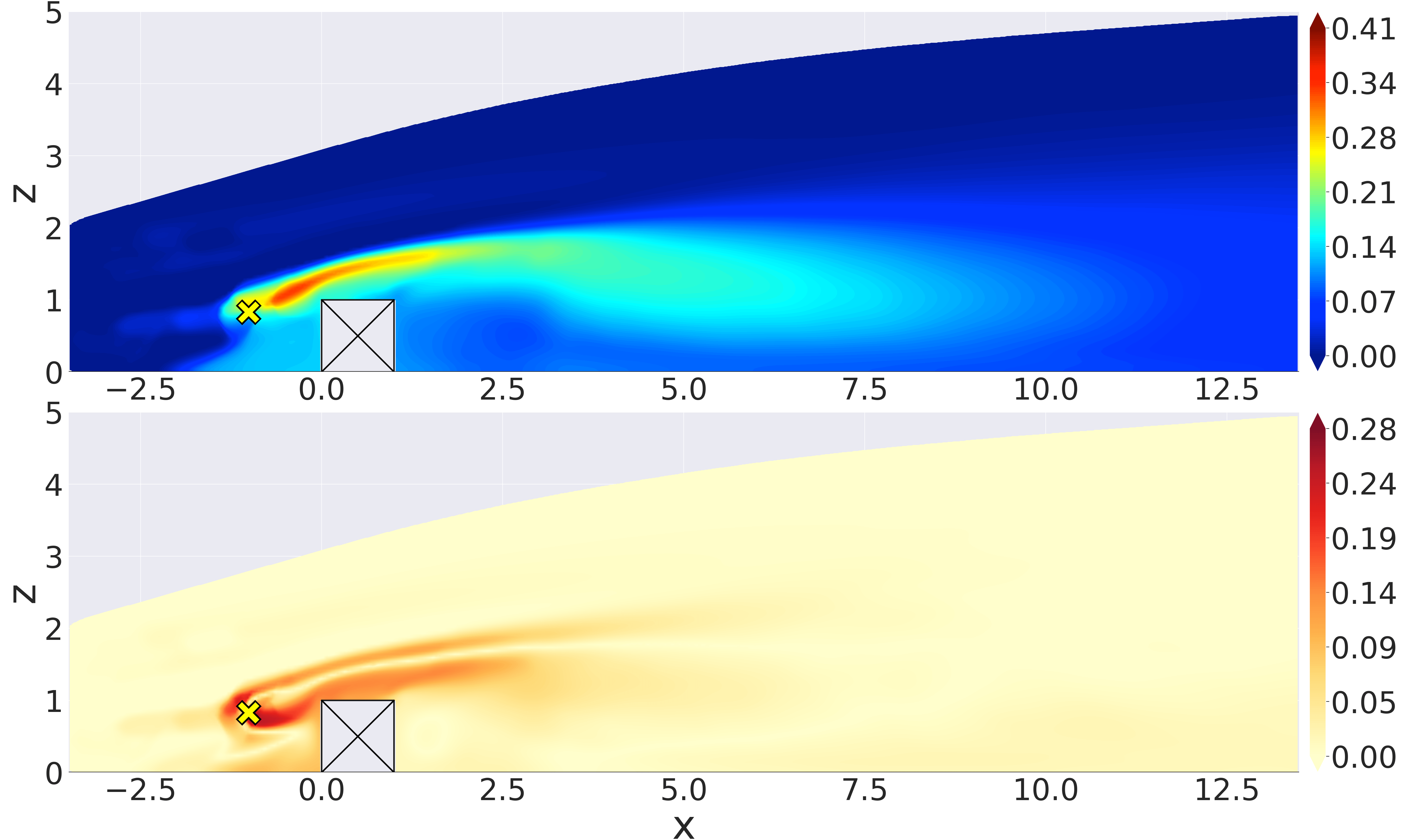}
\caption{Same caption as in Fig.~\ref{fig:nominal_concat} for reduced-order model prediction with $N_{\text{train}} = 100$ training snapshots.}
\label{fig:nominal_robustness}
\end{figure}

Including higher-order modes in the reduced-order model can even degrade its global performance. Here, the optimal choice for $N_{\text{train}} = 100$ is to keep $L = 42$ modes and $L = 25$ modes for $N_{\text{train}} = 50$. The associated global $Q^2$ score is equal to 87.3\% and 85.1\%, respectively, and can be compared to 96.8\% for the full training dataset (Sect.~\ref{sec:res_gp}). These scores may seem satisfactory but when looking at the nominal snapshot prediction, the prediction errors observed before are amplified. Figure~\ref{fig:nominal_robustness} presents the prediction result for $N_{\text{train}} = 100$. The shape of the tracer concentration wake is retrieved but the reduced-order model has difficulty to predict the correct tracer concentration magnitude: the tracer concentration is underestimated at the emission source and upstream of the obstacle, and the high tracer concentrations correspond to a much thinner region than for the full training dataset (Fig.~\ref{fig:nominal_concat}). 

This highlights that the analysis of the overall $Q^2$ score can be misleading about the reduced-order model accuracy, since the prediction quality is spatially heterogeneous and can decrease upstream of the obstacle due to the large tracer concentration gradients near the emission source. Even if the prediction performance is reduced when the training dataset includes fifty to hundred snapshots, the reduced-order model predictions remain physically-consistent. This is no longer the case when further reducing the training dataset (not shown), since some non-physical large-scale structures appear in the predicted tracer concentration fields. Hence, the minimum budget required to emulate the parameterized LES seem to be at least fifty to hundred snapshots.

\subsection{Application to turbulent scalar fluxes}
\label{sec:res_fluctu}

To better exploit the explicit resolution of turbulence by  LES approach (compared to other approaches such as RANS for instance), it is possible to rely on a similar reduced-order modeling approach for higher-order statistical quantities. In the context of dispersion, the turbulent scalar flux is a key quantity that is fully modelled in low-fidelity dispersion models~\citep{vervecken2015rom}. From LES data, reduced-order model for this type of quantities can be extracted. This illustrated in this section by extending our reduced-order modeling approach to the resolved vertical turbulent scalar flux defined and normalized as $(\overline{v K}-\overline{v}\overline{K}) \left(\frac{H^2}{Q_s}\right)$, where $v$ is the vertical component of the velocity.
This demonstration is carried out with the full training dataset ($N_{\text{train}} = 472$) as in Sect.~\ref{sec:res_pod} to Sect.~\ref{sec:res_gp}.

The finer structures make the prediction of these fields slightly more complex than for the mean quantities of interest. The POD variance is indeed spread over a larger number of POD modes: one hundred modes carry 99.3\% of the ensemble variance for the vertical turbulent scalar flux (compared to 99.6\% for the mean quantities of interest); and the Kaiser rule (Sect.~\ref{sec:res_pod}) suggests to keep around 32 modes instead of 25 modes for the mean quantities of interest.

\begin{figure}
\centering
\begin{overpic}[width=0.9\textwidth]{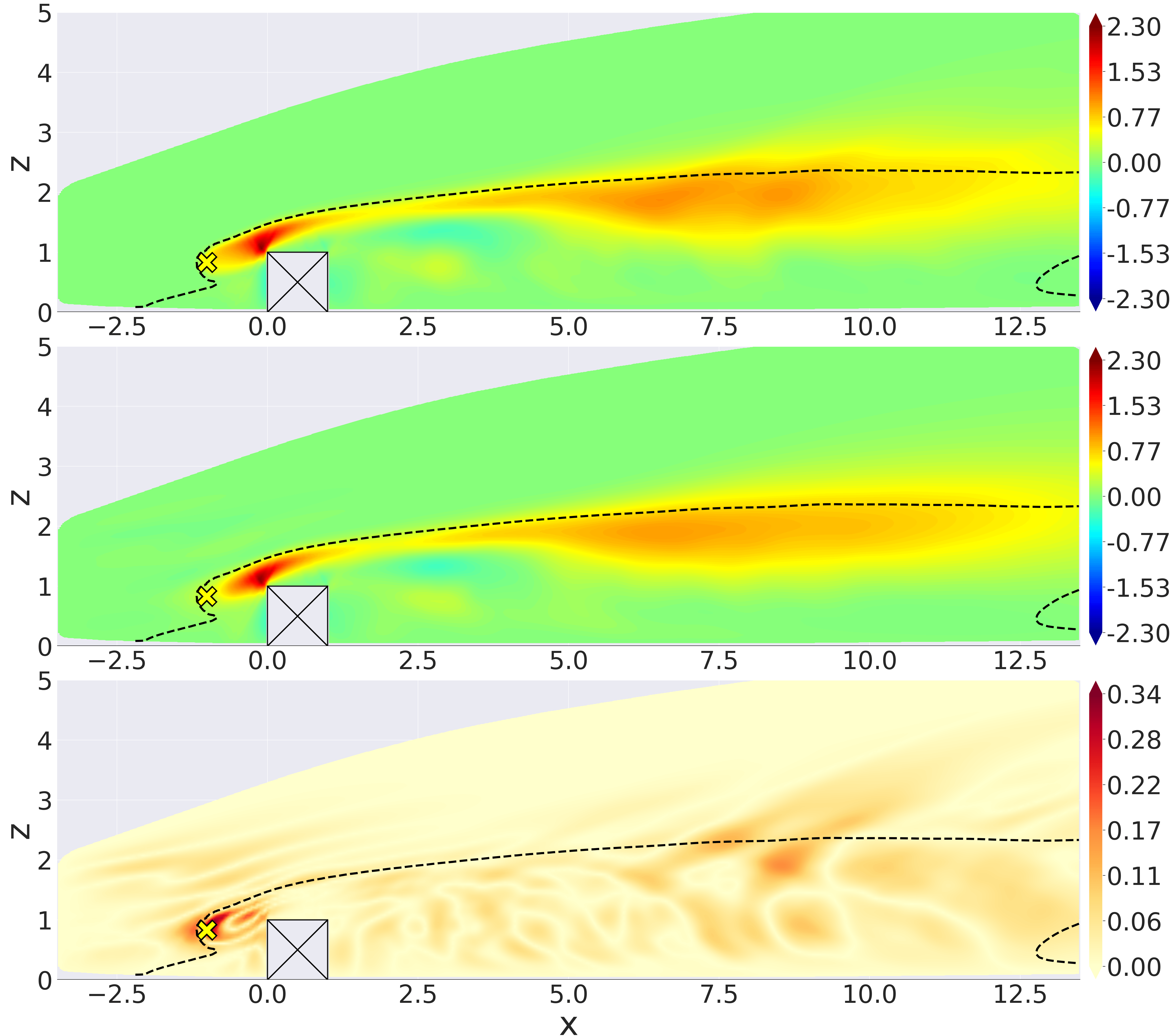}
 \put (7.5,81.5) {\large(a)}
 \put (7.5,53) {\large(b)}
 \put (7.5,24.5) {\large(c)}
\end{overpic}
\caption{Nominal snapshot vertical turbulent scalar flux. (a)~Vertical turbulent scalar flux obtained with LES. (b)~Reduced-order model prediction. (c)~Prediction absolute error. The dashed lines represent the contour line along which the mean normalized tracer concentration is equal to one tenth of the maximum concentration to highlight the area of high concentration values.}
\label{fig:nominal_VK}
\end{figure}

As for the mean quantities of interest, the maximum performance on the test dataset is obtained for $L=100$ modes, and achieves $Q^2 = 94.0\%$ globally for the vertical turbulent scalar flux (compared to 96.8\% for the mean quantities of interest). Hyperparameter trends (not shown) are similar to what we described in Sect.~\ref{sec:res_gp}. The lower $Q^2$ performance can be partly explained by the wider spread of information on the high POD modes. As an illustration, Fig.~\ref{fig:nominal_VK} shows the vertical turbulent scalar flux prediction for the nominal snapshot. We can observe the sharper localized structure compared to the associated mean concentration field (Fig.~\ref{fig:nominal_concat}a), with upward (positive) turbulent scalar flux in a large part of the plume and downward (negative) flux in the wake of the obstacle. Regarding the reduced-order model prediction (Fig.~\ref{fig:nominal_concat}b), the worst areas of predictions are again located upstream of the obstacle close to the source. Still, the sign and magnitude of the turbulent flux is well predicted by the reduced-order model, even in localized areas of strong dispersion, such as the windward corner of the obstacle. This highlights that our reduced-order modeling approach is versatile and has the capacity to emulate a variety of LES field statistics depending on the context of microscale air pollutant assessment: it can perform direct prediction of the tracer concentration, or it can be used to extract physical quantities related to the phenomena that are directly relevant for the plume spread modeling such as turbulent scalar fluxes.

\section{Conclusion}

The goal of this study was to design and validate a non-intrusive reduced-order model combining POD and GPR for predicting LES field statistics of interest associated with near-source tracer concentration dispersion in a multi-query uncertainty quantification context. A two-dimensional case study corresponding to a turbulent atmospheric flow over a surface-mounted obstacle was considered to generate a large LES ensemble database based on perturbed inflow boundary conditions and emission source location.

POD analysis revealed that low-order modes carry concentration information in widely-spread areas and near the obstacle, whereas high-order modes are necessary to include in the POD basis to characterize the concentration high variability near the emission sources. Hundred modes were therefore retained in the reduced-order model, with a mode-per-mode optimization of the Gaussian process hyperparameters to adapt to the wide range of spatial scales across the POD basis.

Two strategies for optimizing GPR hyperparameters were compared. We showed that a satisfactory prior distribution for the hyperparameters can be obtained by analyzing POD modes and used to inform a MAP optimization procedure. MAP was found to provide similar results to a standard N-restart MLL maximization approach from a single gradient descent, providing a reliable efficient offline training framework. This is all the more important as the hyperparameter optimization procedure must be carried out for each POD reduced coefficient.

During the validation stage, we obtained very good global $Q^2$ scores (above 95\%). However, this hides disparities between prediction areas. We demonstrated through a mode-per-mode $Q^2$ analysis that concentration levels in the recirculation areas are well predicted but that concentration levels near the emission source are more challenging to predict since information is carried by high-order POD modes. These high-order modes feature very localized structures associated with perturbed emission location, which are difficult to predict and which are prone to more noise than low-order modes. We also showed that this difficulty of predicting high-order modes increased when the training dataset is reduced, but the prediction performance remains acceptable when considering at least fifty to hundred LES snapshots in the training database. For all these issues, the mode-per-mode analysis was essential to understand the behavior of the reduced-order model.     

Future work includes testing alternative compression approaches to POD. Since POD is based on linear algebra, decomposing information with substantial nonlinearity might be tricky. Deep-learning-inspired techniques such as autoencoders offer an intriguing line of study for improving compression performance~\citep{fukami2020convolutional} and for meeting the constraint of reduced training database. From an application viewpoint, future work includes extending the reduced-order modeling approach to more realistic atmospheric boundary-layer flows to evaluate its potential for air pollutant dispersion applications. In the long term, a reduced-order model informed by LES could be valuable to produce ensemble forecasts and assess human exposure to toxic air pollutants in the event of an accident. 

\section*{Statements \& Declarations} 

\paragraph{Funding}
Funding and access to supercomputing resources for this work were provided by CERFACS. 

\paragraph{Competing Interests}
All authors have no competing interests to declare that are relevant to the content of this article.

\section*{Acknowledgements}
The authors acknowledge Thibault Gioud (Cerfacs) for support on the LEMMINGS Python package for managing ensemble LES runs. They also acknowledge Isabelle D'Ast from the Computer Support Group at Cerfacs for support on GPyTorch and TensorFlow library installations.

\bibliographystyle{abbrvnat}  
\bibliography{references}

\end{document}